\documentclass{article}

\pdfoutput=1

\usepackage{arxiv}

\usepackage[utf8]{inputenc} % allow utf-8 input
\usepackage[T1]{fontenc}    % use 8-bit T1 fonts
\usepackage{hyperref}       % hyperlinks
\usepackage{url}            % simple URL typesetting
\usepackage{booktabs}       % professional-quality tables
\usepackage{amsfonts}       % blackboard math symbols
\usepackage{nicefrac}       % compact symbols for 1/2, etc.
\usepackage{microtype}      % microtypography
\usepackage{lipsum}
\usepackage{graphicx}
\graphicspath{ {./Assets/} }

\usepackage[numbers]{natbib}
\usepackage{todonotes}
\usepackage{amsmath,amsfonts,amssymb}
\usepackage{bm}
\usepackage{graphicx, rotating}
\usepackage{float}
\usepackage{textcomp}
\usepackage{multirow}
\usepackage{arydshln}

\graphicspath{ {./Assets/} }

\newcommand{\Real}{\mathbb{R}}
\newcommand{\real}{\Real}
\newcommand{\E}[2]{\mathbb{E}_{#1}{[#2]}}

\usepackage{xspace} 
\makeatletter
\DeclareRobustCommand\onedot{\futurelet\@let@token\@onedot}
\def\@onedot{\ifx\@let@token.\else.\null\fi\xspace}

\def\eg{\emph{e.g}\onedot} 
\def\ie{\emph{i.e}\onedot}

\def\wrt{w.r.t\onedot}

\newcommand{\fr}[1]{\textcolor{red}{$^\diamond$}}

\usepackage{soul}

\title{On the Transferability of VAE Embeddings using Relational Knowledge with Semi-Supervision}

\author{
 Harald Str\"omfelt \\
  Department of Computing \\
  Imperial College London \\ 
  London, SW7 2AZ \\
  \texttt{h.stromfelt17@imperial.ac.uk} \\
   \And
 Luke Dickens \\
  Department of Information Studies\\
  University College London\\
  London, WC1E 6BT \\
  \texttt{l.dickens@ucl.ac.uk} \\
  \And
 Artur d'Avila Garcez \\
  Department of Computer Science \\
  City University of London \\
  London, EC1V 0HB \\
  a.garcez@city.ac.uk \\
  \And
 Alessandra Russo \\
  Department of Computing \\
  Imperial College London \\ 
  London, SW7 2AZ \\
  \texttt{a.russo@imperial.ac.uk} \\
}

\begin{document}
\maketitle

\begin{abstract}
%Semi-supervised Variational AutoEncoders (VAE) have had success in obtaining data-representations that expose their semantic factors. 
We propose a new model for relational VAE semi-supervision capable of balancing disentanglement and low complexity modelling of relations with different symbolic properties.
We compare the relative benefits of relation-decoder complexity and latent space structure on both inductive and transductive transfer learning.  
Our results depict a complex picture where enforcing structure on semi-supervised representations can greatly improve zero-shot transductive transfer, but may be less favourable or even impact negatively the capacity for inductive transfer.
\end{abstract}

\section{Introduction}
\label{sec:introduction}
When dealing with complex data, the effectiveness of a classifier/predictor is limited by its ability to extract useful information. As such, representations that clearly expose the semantics of the data should then be most amenable to downstream learning \cite{Pan2009-ASO, Bengio2013-RLR}.
This is often referred to as a challenge of acquiring a \textit{disentangled} representation over the factors of the data 
\cite{Higgins2017-BVL}.
A popular recent trend that has had significant success in this regard uses semi-supervised Variational AutoEncoders (VAE) 
\cite{ Shu2020-WSD, Locatello2020-WSD, Chen2019-WSD, Karaletsos2016-WCH, Kingma2014-SSL, Feng2018-DSD}.
Whilst fully unsupervised VAE methods have been shown to require strong inductive bias \cite{Locatello2019-CCA}, semi-supervised methods achieve disentanglement by training additional auxiliary tasks that are defined on the factors, alongside the standard VAE objective (see Appendix Eqn. \ref{eq:elbo}). 

% The problem and what we do
Recently relation-learning as semi-supervision to VAE representation learning has shown promise in shaping the representations learned \cite{Karaletsos2016-WCH, Chen2019-WSD, Chen2019-ROV}.
In practice, different relations between data are often interrelated if they are derived from shared underlying factors. We argue
that this presents a trade-off between decoder complexity accuracy achievable via highly complex
decoders and the value of a latent representation carries over to new data or tasks. As simpler
decoders capture fewer independent relationships, they can provide a structural bias towards a
beneficial sharing of semantic factors. However, overly simple decoders may only be able to express
some global properties of relations and not others, e.g. symmetry, transitivity, etc.
We explore this trade-off by investigating the inductive and transductive transfer performance of two relation-decoders: the ``Neural Tensor Network'' (NTN), a powerful latent factor model (LFM) \cite{Trouillon2019-OIA, Nickel2016-ARO, Wang2017-KGE, Socher2013-RWN}; and our own novel \emph{Dynamic Comparator} (DC) model with 10$\times$ fewer parameters. While our DC decoder has stricter constraints on the expected latent space structure 
than NTN, it is still sufficiently flexible to express a broad class of global properties on
relations. We evaluate these ideas on a variety of tasks using MNIST digit images, our results show that: 1. semi-supervision improves inductive transfer by an appreciable margin (also seen in \cite{Locatello2019-CCA}). 2. Strongly structuring the latent space can degrade the inductive transfer capacity of encodings; and 3. Over 90\% zero-shot transductive transfer accuracy on a binary task with semi-supervised VAE representations using our proposed DC model, significantly outperforming no semi-supervision (< 76\%) and NTN based semi-supervision (< 80\%).

In the following, Section \ref{sec:model} introduces the model, Section \ref{sec:experiments} presents and then discusses the experimental results. We include additional background on LFM and VAE in Appendix \ref{apdx:sec:background_theory}.

\begin{figure}
    \centering
    \includegraphics[width=1.\textwidth]{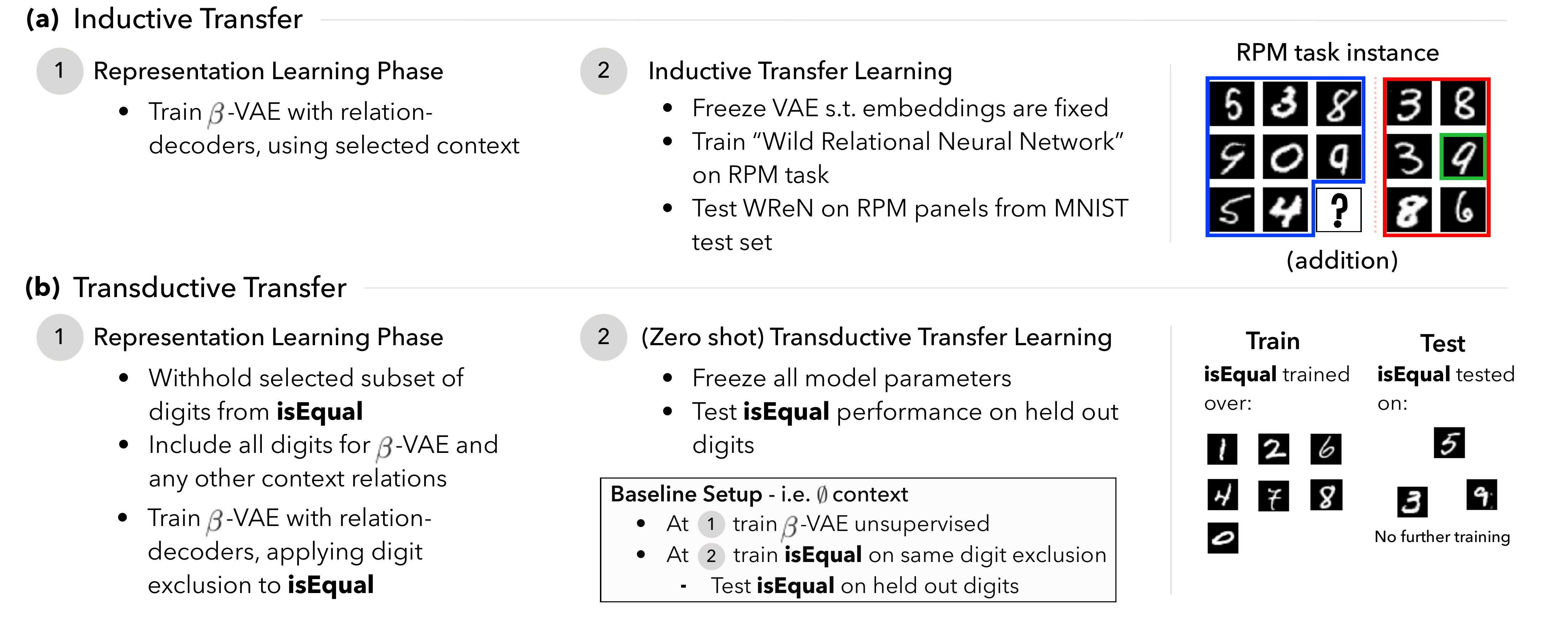}
    \caption{Overview of the transductive and inductive experimental setup used in the evaluation.}
    \label{fig:experimental_setup}
\end{figure}
\section{A simple but flexible relation-decoder}
\label{sec:model}
In this paper, we focus our work on \textit{MNIST} \cite{mnist} and incorporate combinations of the $\{\textsf{isEqual}, \textsf{isGreater}, \textsf{isSuccessor}\}$ binary numeric relations, in order to disentangle the `number' factor. 
However, as each can have different symbolic properties (\eg symmetry, transitivity and so on) % (see Table \ref{tab:relations_symbolic_properties}).
we are presented with possible trade-off between model flexibility, in terms of the types of relations they can learn, and the degree to which they enforce disentanglement.
On the one hand, restrictive relation-decoders that enforce disentanglement may not be able to model each relation type. For instance, \citet{Chen2019-WSD} was able to disentangle digit identity on MNIST using the $\textsf{isEqual}$ relation - however, the proposed relation-decoder cannot model asymmetrical relations. In contrast, higher capacity decoders may generalise to complex relations but at the expense of instead negatively affecting disentanglement.
% In the following we present a new relation-decoder designed to model a wider range of relations whilst enforcing disentanglement. We then explore the relative transfer benefits of choosing more or less structured relation-decoders.

% \subsection{Dynamic Comparator}
To bridge the gap, we propose the following ``\textbf{D}ynamic \textbf{C}omparator'' (\textbf{DC}) model, a new LFM that encourages disentanglement whilst being able to model (a)symmetric and (non-)transitive relations:
\begin{equation}
    f^{DC}_r(\bm{z}_i, \bm{z}_j) = 
    a_0 \cdot \underbrace{
    \sigma\big(\eta_0( \eta_1 - \|\bm{u}\odot (\bm{z}_i - \bm{z}_j + \bm{b}_\dagger)\|_2^2)\big)
    }_{f_r^\dagger}
    + 
    a_1 \cdot \underbrace{
    \sigma\big((\eta_2 \cdot \bm{u}^\top (\bm{z}_i - \bm{z}_j + \bm{b}_\ddagger))\big)
    }_{f_r^\ddagger}.
    \label{eq:lfm*}
\end{equation}
For relation $r$, given $m$-dimensional latent representations $\bm{z}_i, \bm{z}_j \in \real^m$ obtained via a VAE encoder: $\bm{a} = \texttt{Softmax}(\bm{A}) \in \real^2$ is an attention weighting between the two functional forms $f_r^\dagger$ and $f_r^\ddagger$; $\bm{u} = \texttt{Softmax}(\bm{U}) \in \real^{m}$ is an attention mask over the full $m$-dimensional latent space; $\bm{b}_\dagger, \bm{b}_\ddagger \in \real^m$ are additional learnable bias terms; and $\eta_0, \eta_1 \in \real^+$ are non-negative and $\eta_2 \in \real$ any-valued scalar terms, respectively. Lastly, $\sigma$ is the sigmoid function used to bound the output to [0,1], $\odot$ denotes element-wise multiplication and $\|\cdot\|_2$ is the $L2$-norm.
% Description of the components
$f_r^\dagger$ is a generalisation of the relation function from \cite{Chen2019-WSD} designed for the $\textsf{isEqual}$ relation and only capable of modelling symmetric `zero-centred' relations.
Firstly, by including $\bm{b}_\dagger$, the $L2$-norm can depend on the order of $\bm{z}_i$ and $\bm{z}_j$, enabling the modelling of asymmetric relations such as $\textsf{isSuccessor}$. Further, whilst \cite{Chen2019-WSD} hard-code the relevant subspace as a hyperparameter, we include a learned mask, $\bm{u}$, which allows the function to `bind' itself to the relevant latent variable such that latent distance is only calculated on this subspace - this approach was previously done in \cite{Karaletsos2016-WCH}. In common with \cite{Chen2019-WSD}, $\eta_0$ sets the steepness of the true/false decision boundary and $\eta_1$ is a threshold that sets the width of the relation.
$f_r^\ddagger$ generalises \cite{Chen2019-ROV} whom omit $\bm{b}_\ddagger$ and set $\eta_2$ to be a non-negative scalar that models confidence in the relation. As such, \cite{Chen2019-ROV} strictly models one-way ordinal `$>$' relations and critically has fixed predictions at equality (\ie $\sigma(0)=0.5$). In contrast, the proposed $f_r^\ddagger$ can learn the ordering of the relation and can learn any of $>, \ge, \le, <$ type relations, such as $\textsf{isGreater}$. Once again, \cite{Chen2019-ROV} hard-code the sub-space for the ordering, whereas \textbf{DC} can discover it via $\bm{u}$ as we perform a dot product with the mask to calculate the directional $\bm{z}_i, \bm{z}_j$ difference on the $\bm{u}$ mask hyperplane.
Importantly, \textbf{DC} can learn combinations of semantically similar relations such that they are each calculated using the same latent factors, which can support disentanglement. See Appendix \ref{apdx:sec:relation_details} for further details.

In the next section, we compare the relative transfer performance obtained when using either \textbf{NTN+} (a modified version of NTN \cite{Donadello2017-LTN, Serafini2016-LTN} for $n$-ary relations - see Appendix Eqn. \ref{eq:ntn+}) or \textbf{DC} relation-decoders.

%NTN+ is also fully expressive, meaning that for a sufficiently large embedding size, it can model any number of true/false triples.

\section{Experiments}
\label{sec:experiments}
In this section we compare the quality of representations produced by VAE semi-supervision using either \textbf{NTN+} or the proposed \textbf{DC}, against those generated by unsupervised $\beta-VAE$, for both inductive and transductive transfer. \cite{Pan2009-ASO} (see Figure \ref{fig:experimental_setup}).
In all experiments, we generate representations for digits from the MNIST dataset and introduce semi-supervision with additional predictions for binary relations isEqual (Eq), isGreater (Gr) and isSuccessor (Su). We consider four configurations of relations for semi-supervision: $\emptyset$ (no supervision), \{Eq\}, \{Eq, Gr\} and \{Eq, Gr, Su\}, and call this the \emph{context}.
Our approach could readily be adapted to work with a variety of unsupervised VAE, including those developed for disentanglement \cite{Chen2018-ISD, Kumar2018-VID, Kim2018-DF, Ridgeway2018-LDD, Eastwood2018-FQE}.  However, we chose to use the $\beta$-VAE due it showing competitive results whilst being straightforward to optimise given that it only has one hyperparameter, $\beta$, which is understood to control disentanglement pressure \cite{Higgins2017-BVL, Burgess2017-UDI}. This leads to the following joint objective,
\begin{align}
\ln p_\theta(\bm{X}, \bm{Z}) \ge \mathcal{L}_{\beta\text{-VAE}}^{ELBO} - \lambda 
\underbrace{
\E{r, \bm{z}_i, \bm{z}_j, y_{ij} \in \mathcal{T}}{
y_{ij} \ln(\hat{y}_{ij}) + (1-y_{ij}) \ln(1-\hat{y}_{ij})
}}_{\mathcal{L}^{LFM}},
\label{eq:objective}
\end{align}
where $\hat{y}_{ij}$ is estimated by $f_r(\bm{z}_i, \bm{z}_j)$ and $\mathcal{T}$ is the set of all positive ($y_{ij} = 1$) and negative triples ($y_{ij} = 0$) of the form $(\bm{z}_i, r, \bm{z_j})$. 
$\mathcal{L}_{\beta\text{-VAE}}^{ELBO}$ is the $\beta$-VAE ELBO (see Appendix \ref{apdx:sec:background_theory}) and $\lambda$ is a weighting parameter.
All latent representations are sampled according to $\bm{z} \sim q_\phi(\bm{Z}| \bm{X})$ where $q_\phi(\bm{Z}| \bm{X})$ is modelled by the VAE encoder.
Given a context, we sample two triples per \textit{MNIST} image and randomly select which relation to generate the triple for - this ensures a fixed number of triples between experiments.
We direct readers to Appendix \ref{apdx:sec:background_theory} for further details on the $\beta$-VAE and \textbf{NTN+}, and Appendix \ref{sec:apdx:model_details} for additional implementation details.

\begin{figure}
    \centering
    \includegraphics[width=1.\textwidth]{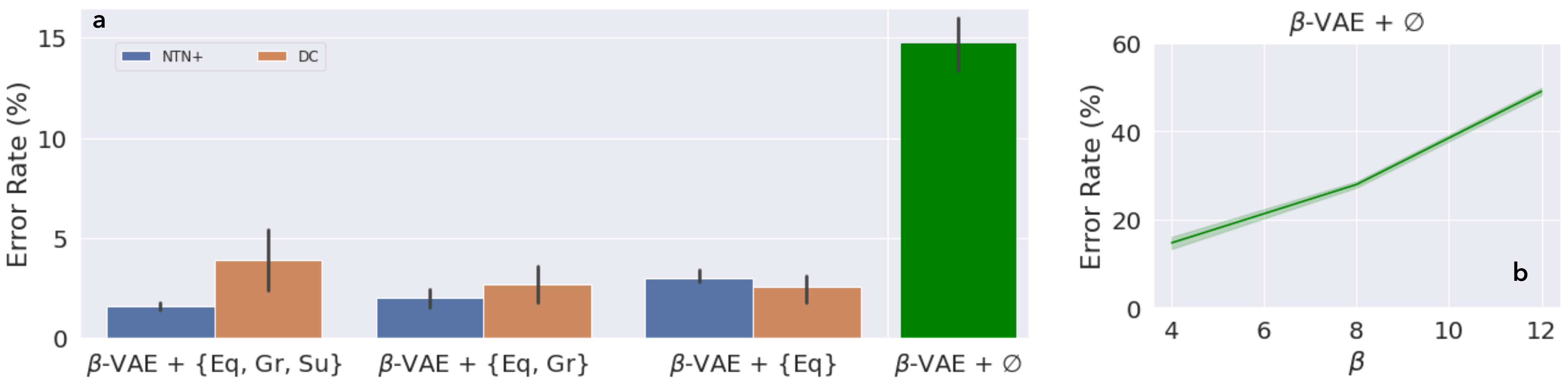}
    \caption{Inductive transfer error rate on our \textit{MNIST} RPM task. (a) demonstrates the performance per context and relation-decoder with $\beta=4$ and (b) demonstrates $\beta$ effects on an unsupervised $\beta$-VAE. We see a clear shift in performance when semi-supervision is used, with marginally worse performance for \textbf{DC}, and negative correlation of performance with $\beta$ increases for $\emptyset$ context.}
    \label{fig:macro_avr_comparison}
\end{figure}

%%%%%%%% INDUCTIVE SECTION %%%%%%%%
\textbf{Inductive Transfer:} In this setting both the source and target data are the same but the target (downstream) task differs \cite{Pan2009-ASO}. 
%%%%%%
We follow recent work \cite{Steenbrugge2018-IGA, Steenkiste2019-ADR, Barrett2018-MAR} and create a RPM dataset consisting of $3\times3$ \textit{MNIST} image panels, arranged into rows of addition or subtraction. The final row is left incomplete and a downstream reasoner is tasked with using the VAE image-encoding to select the correct tile (see the addition example in Figure \ref{fig:experimental_setup}(a)-right). We explore the downstream performance effects that different forms of semi-supervision have. 
see Appendix \ref{apdx:sec:avr} for a detailed description of the RPM task and downstream reasoner.
If digit classification is possible, it would be possible to complete this task by memorizing the addition/subtraction combinations, however including numeric ordering should alleviate the need for memorization.
The aim of this experiment is to evaluate the inductive transfer improvement that semi-supervision produces and the relative benefit of regularising for further structure beyond digit identity.
%%%%%%
\textbf{Results:} 
Figure \ref{fig:macro_avr_comparison}(a) shows the maximum 5000-step moving average test error rate obtained using each context and relation-decoder and (b) demonstrates how $\beta$ settings affect the downstream performance using an unsupervised VAE. 
%%%%%
\textbf{Discussion:} 
These results are in agreement with recent work that showed semi-supervision improves inductive feature-representation-transfer \cite{Locatello2019-CCA, Chen2019-WSD}. However, in contrast with \cite{Steenbrugge2018-IGA, Steenkiste2019-ADR}, we observe a negative correlation with $\beta$ increase (Figure \ref{fig:macro_avr_comparison}(b)).
In conjunction with an increasing error rate as more relations are included for \textbf{DC} but not for \textbf{NTN+} (Figure \ref{fig:macro_avr_comparison}(a)), the results indicate that enforcing stronger regularity in the latent embeddings worsens performance.
In support of this, Appendix Figure \ref{fig:best_vs_worst_inductive_DC} shows that adding more context increases the overall digit factor information captured using \textbf{NTN+}. In summary, reduced inductive transfer performance may occur when digit class identity is obscured.

\begin{figure}
    \centering
    \includegraphics[width=1.\textwidth]{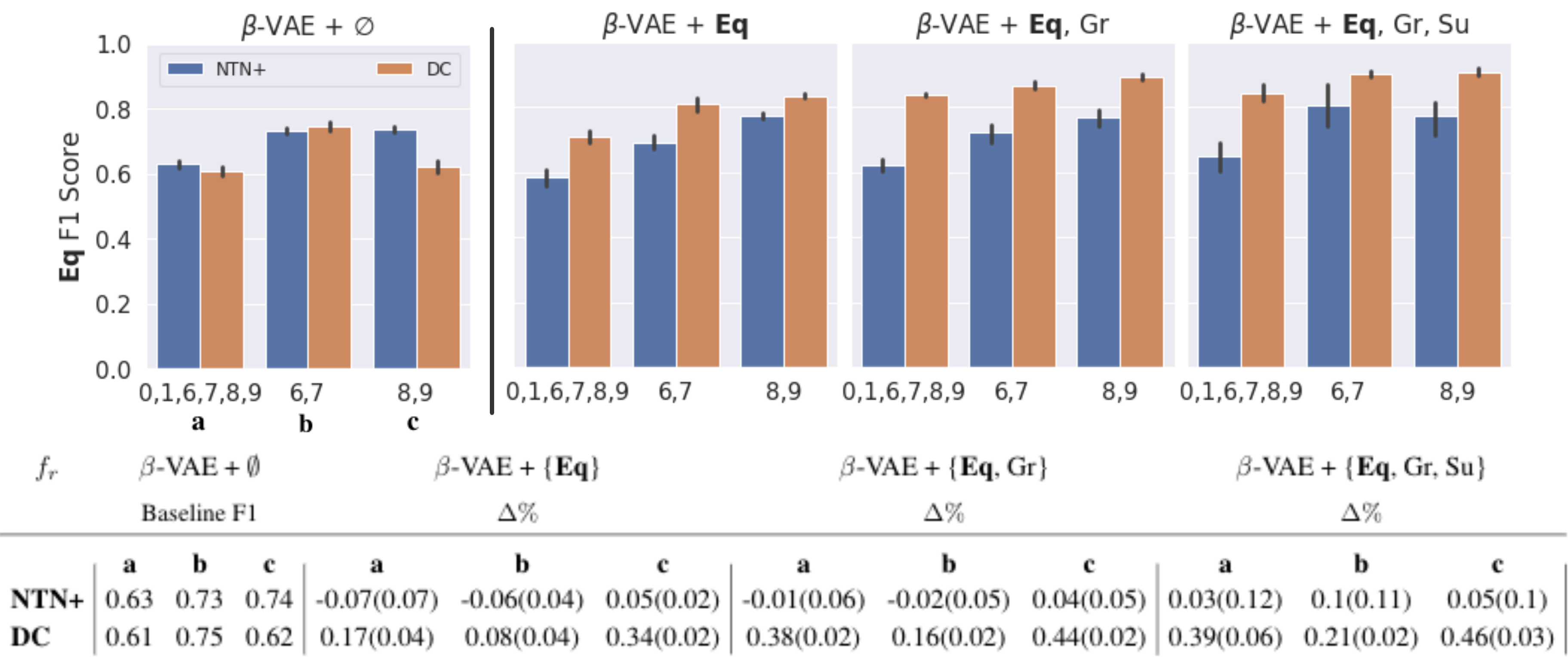}
    \caption{Zero-shot transductive transfer results showing (\textbf{upper}) the $\textsf{isEqual}$ F1 scores on the \textit{held out} subset digit classes (indicated by the horizontal axis groupings); and (\textbf{lower}) $\Delta\%$ difference \wrt baseline F1 for each relation-decoder and context, with standard deviation in parentheses. Both results obtained with $\beta=4$ for the VAE.
    These results indicate that lower complexity decoders perform better at transductive transfer.}
    \label{fig:macro_excludedclasses_comparison}
\end{figure}

%%%%%%%% TRANSDUCTIVE SECTION %%%%%%%%
\textbf{Transductive Transfer:} In contrast to inductive transfer, here the source and target task data is different but the task itself, in this case \textsf{isEqual} relation prediction, is equivalent in both cases. 
%%%%%%
Concretely, we only train the \textsf{isEqual} relation on a subset of the data by omitting a selection of digits, but show all digits to the VAE and other relations. 
Hence, in the source domain \textsf{isEqual} only observes a subset of the digits and is then tested on the unseen digits, wherein no further training takes place; as such we test \emph{zero-shot} transductive transfer.
With this experiment, we aim to evaluate the amenability of the representations obtained using different contexts and relation-decoders, to the \textsf{isEqual} relation-decoder parameterization learned on the digit subset. 
%%%%%%
\textbf{Results:} 
Figure \ref{fig:macro_excludedclasses_comparison} compares the \textsf{isEqual} F1 test scores on the held out digits, when learned using each relation-decoder. We use $\emptyset$ context as a baseline, wherein we pre-train an unsupervised $\beta$-VAE and post-train \textsf{isEqual} using each relation-decoder on frozen embeddings, with the same digit exclusion strategy.
%%%%%
\textbf{Discussion:} 
in the baseline case, each relation-decoder cannot influence the VAE-encodings and so must `fit' to the frozen latent embeddings that result from the pre-trained $\beta$-VAE. Interestingly, both decoders perform similarly even though they have different complexities.
We then observe marked performance increases when including \textsf{Eq} for \textbf{DC} but not \textbf{NTN+}. This suggests that \textbf{DC} is able to impose stronger regularity on the $\beta$-VAE such that the resulting latent embeddings exhibit regularity that is amenable to \textbf{DC}, even on untrained digits.
As expected, we observe a significant improvement for \textbf{DC} when including \textsf{Gr} since it observes the full digit set, for example improving by 39\% over the baseline for exclusion setting (\textbf{a}). However, \textbf{NTN+} does not exhibit the same improvement. This may indicate that, although $\textsf{Gr}$ is symbolically related to $\textsf{Eq}$, it may not be learned in such a way that this relatedness is captured between the latent embeddings and the \textsf{Eq} versus \textsf{Gr} relation-decoder parameters.
Lastly, the increased variance for \textbf{NTN+} on context \{Eq, Gr, Su\} is likely due to it requiring more data to be trained - since we use a fixed triple ``quota'' that is shared between relation-decoders, adding more relations reduces the total number of triples observed by each relation-decoder. \textbf{DC} is by contrast more data efficient, due to it having far fewer parameters to learn.
In summary, \textbf{DC} outperforms both the baseline and \textbf{NTN+} in each exclusion setting, with immediate gains for \{Eq\} context wherein no relations are trained on the held out digits. This indicates that better transductive transfer performances can be achieved when using relation-decoders that can impose consistent regularity on the $\beta$-VAE with respect to the relation-decoder parameters.

\textbf{Concluding remarks:} 
The results in this paper shed light onto the complex interplay between latent embedding structure and the decoders that are used to perform downstream tasks. In order to obtain transferable latent representations, we observe that for powerful neural network based downstream learners, stronger regularity is less favourable. On the other hand, our results suggest that we can achieve better transductive transfer results if we enforce regularisation on the representations. This has the potential of encouraging a consistent structure across the latent space which relation-decoders can leverage.

\newpage

% BIB

\newpage

%% Appendix
\appendix

\section{Relation-decoder case study}
\label{apdx:sec:relation_details}
In this section, we expose what is learned, in terms of the parameterization and resulting latent embeddings, when training each relation-decoder alongside a $\beta$-VAE. 
For DC, we include mask visualisations ($\bm{u}$ in Eqn. \ref{eq:lfm*}) to examine which latent subspace is used to calculate each relation, as well as the full parameterization of DC for \textit{MNIST}.
We then include an exploration of each latent dimension against ground truth factors for both \textit{MNIST} and the benchmark \textit{dSprites} dataset, which consists of grey-scale images of hearts, square and ovals; each varying across scale, orientation and position \cite{Higgins2017-BVL}
For both datasets, Mutual Information Gap (MIG) scores are calculated - the MIG score calculates the normalised difference between the two latent dimensions that share the greatest mutual information \wrt each ground truth factor \cite{Chen2018-ISD}. When quoted as a single value it is averaged across ground truth factors:
\begin{align}
    \frac{1}{K} \sum_{k=1}^{K} \frac{1}{H(v_k)}\big(I_i(z_{j^{(k)}} ; v_k) - \max_{j \neq j^{(k)}} I_i(z_j ; v_k)\big) \nonumber
\end{align}
where $K$ is the total number of ground truth factors, $I_i(\cdot ; \cdot)$ is the mutual information between two random variables for input $\bm{x}^i$, $H(\cdot)$ is the entropy of a given random variable which acts as a normalisation term and $z_j^{(k)}$ is the latent factor that maximises the mutual information with factor $v_k$. Note that the $k$th factor gives a large contribution, if one dimension of $\bm{z}$, say $z_{j(k)}$, can explain a large amount of the variation in $v_k$, while other dimensions explain little.  
For \textit{MNIST}, we use the `digit' factor (K=1) and for \textit{dSprites}, we follow \cite{Chen2018-ISD} and present the MIG score as an average over {scale, orientation, y-position, x-position} factors ($K=5$).

\subsection{Dynamic Comparator}
\begin{table}
  \caption{Parameters learned for each \textit{MNIST} relation included for representation learning. Parameter names refer to Eqn. \ref{eq:lfm*}. In particular, $\bm{a} = [a_0\quad a_1]$ weightings indicate which functional form is used to model the relation. These values are complimented by the relation function parameter visualisation given by Figure \ref{fig:dc_parameter_visualisation}.}
  \label{tab:dc_mnist_parameters}
  \centering
        \begin{tabular} {l|c|c}
        \toprule
         \addlinespace[0.15em]
         \textbf{Relation} & \textbf{Param.} & \textbf{Value}\\
         \addlinespace[0.15em]
         \midrule
isEqual & $ \eta_2 $ & [-0.]\\
isEqual & $ \mathbf{b_\ddagger} $ & [-2.31]\\
isEqual & $ \mathbf{b_\dagger} $ & [-0.]\\
isEqual & $ \eta_0 $ & [190]\\
isEqual & $ \eta_1 $ & [-0.04]\\
isEqual & $ \mathbf{a} $ & [0.01 0.99]\\
isGreaterThan & $ \eta_2 $ & [70]\\
isGreaterThan & $ \mathbf{b_\ddagger} $ & [-0.18]\\
isGreaterThan & $ \mathbf{b_\dagger} $ & [0.38]\\
isGreaterThan & $ \eta_0 $ & [0.]\\
isGreaterThan & $ \eta_1 $ & [2.24]\\
isGreaterThan & $ \mathbf{a} $ & [0.99 0.01]\\
isSuccessor & $ \eta_2 $ & [-100]\\
isSuccessor & $ \mathbf{b_\ddagger} $ & [-2.65]\\
isSuccessor & $ \mathbf{b_\dagger} $ & [-0.35]\\
isSuccessor & $ \eta_0 $ & [170]\\
isSuccessor & $ \eta_1 $ & [1.17]\\
isSuccessor & $ \mathbf{a} $ & [0.01 0.99]\\
         \addlinespace[0.15em]
        \bottomrule
    \end{tabular}
 \end{table}
 % Table

\begin{figure}
    \centering
    \includegraphics[width=0.8\textwidth]{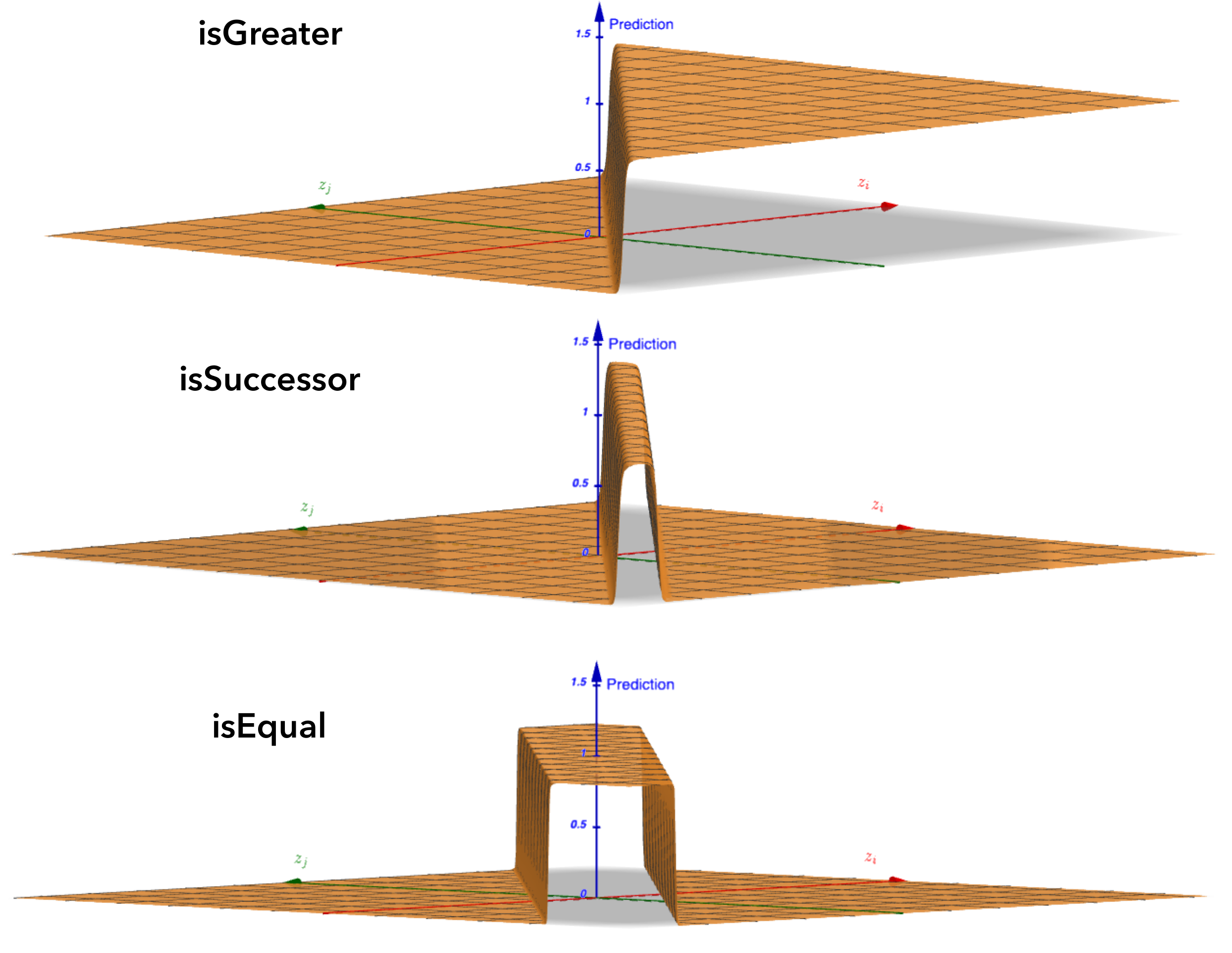}
    \caption{Visualisation of \textbf{DC} parameters given by Table \ref{tab:dc_mnist_parameters}. $x$- and $y$-axis (red and green) correspond with $\bm{z}^6_i$ and $\bm{z}^6_j$ respectively, whilst the $z$-axis (blue) gives the relation-decoder output. We can see that \textsf{isEqual} is learned as a symmetric function, whilst  \textsf{isGreater} and \textsf{isSuccessor} are asymmetric. For \textsf{isGreater}, we can see that non-zero $\bm{b}_{\ddagger}$ ensures that  $f^{DC}_{\textsf{isGreater}}(\bm{z}^6_i, \bm{z}^6_j) \approx 0 $ if $\bm{z}^6_i = \bm{z}^6_j$. In terms of transitivity, we can see that \textsf{isGreater} is globally transitive whilst \textsf{isEqual} and \textsf{isSuccessor} can only model localised forms of transitivity.}
    \label{fig:dc_parameter_visualisation}
\end{figure} 

\begin{figure}
    \centering
    \includegraphics[width=.99\textwidth]{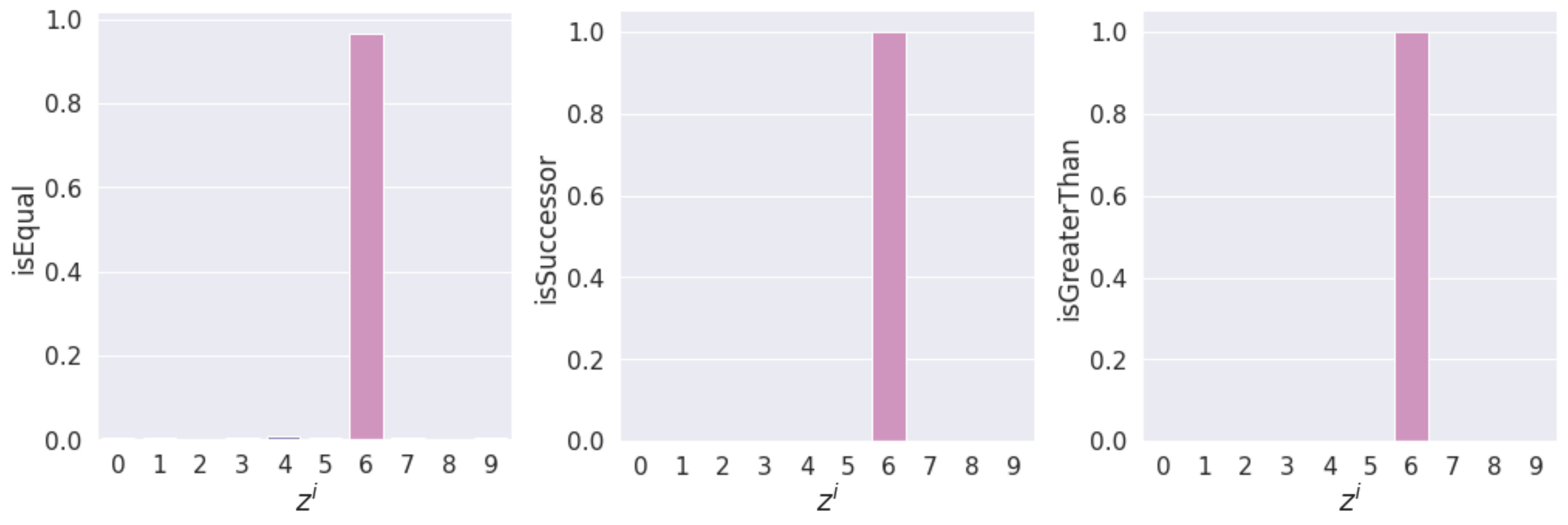}
    \caption{Example of the $\bm{u} = \texttt{Softmax}(\bm{U})$ masks learned by \textbf{DC} on \textit{MNIST} for context $\beta$-VAE + \{Eq, Gr, Su\}. For each relation, \textbf{DC} relation-decoders end up using the same latent subspace, namely $z^6$.}
    \label{fig:MASK_visual_mnist}
\end{figure}

To begin with, Table \ref{tab:dc_mnist_parameters} presents an example of the learned DC parameters after training $\beta$-VAE + \{Eq, Gr, Su\} on \textit{MNIST}.
Figure \ref{fig:dc_parameter_visualisation} then includes true/false region visualisation for each relation-decoder to enable visualisation of the function itself - this is possible since the masks (shown by Figure \ref{fig:MASK_visual_mnist}) select 1-dimensional subspaces, namely $\bm{z}^l$ with $l=6$, so we can plot the relation-decoder output over $\bm{z}^6_i$ versus $\bm{z}^6_j$.

As shown by Figure \ref{fig:dc_parameter_visualisation}, each of \textsf{isEqual, isGreater} and \textsf{isSuccessor} are modelled differently, in accordance with their dissimilar symbolic properties (see Table \ref{tab:relations_symbolic_properties}).
Firstly, the transitive and asymmetrical \textsf{isGreater} relation uses  $f^\ddagger_r$ and is thus modelled by a step function around $(\bm{z}^6_i - \bm{z}^6_j) > 0$. Here, a non-zero $\bm{b}_\ddagger$ ensures that  $f^{DC}_{\textsf{isGreater}}(\bm{z}^6_i, \bm{z}^6_j) \approx 0 $ if $\bm{z}^6_i = \bm{z}^6_j$. 
\textsf{isSuccessor}, which is non-transitive and asymmetrical, is modelled as a relative distance based function, through the use of $f^\dagger$. However, unlike \textsf{isEqual}, which is by contrast symmetric and thus invariant to input ordering, \textsf{isSuccessor} includes an offset offset via $\bm{b}_\dagger \neq 0$ and sets a narrow channel-width via $\eta_1$. This leads to $\textsf{isSuccessor}(\bm{z}^6_i, \bm{z}^6_j) \implies \textsf{isEqual}(\bm{z}^6_i, \bm{z}^6_j - \bm{b}_\dagger)$.
For each relation, decision thresholds are set to be steep, using $\eta_0$ for \textsf{isEqual} and \textsf{isSuccessor}, and $\eta_2$ for \textsf{isGreater}.

An important nuance between each relation is their strictness over transitivity. Whilst $f^\ddagger_r$ will produce `global' transitivity since it outputs 1 for all $\bm{z}^6_i > \bm{z}^6_k$. On the other hand, by incorporating a distance measure, it is important that all triples that are true under $f_r^\dagger$ will maintain a maximum distance between head and tail within the $\eta_1$ channel width. It therefore makes sense that \textsf{isSuccessor} learns a small $\eta_1$ - if it was larger, it is possible that \textsf{isSuccessor} would demonstrate transitivity across two digits. We refer to this as `localised-transitivity' since transitivity will depend on the distances between inputs included in any transitive clause.

\begin{figure}
    \centering
    \includegraphics[width=.99\textwidth]{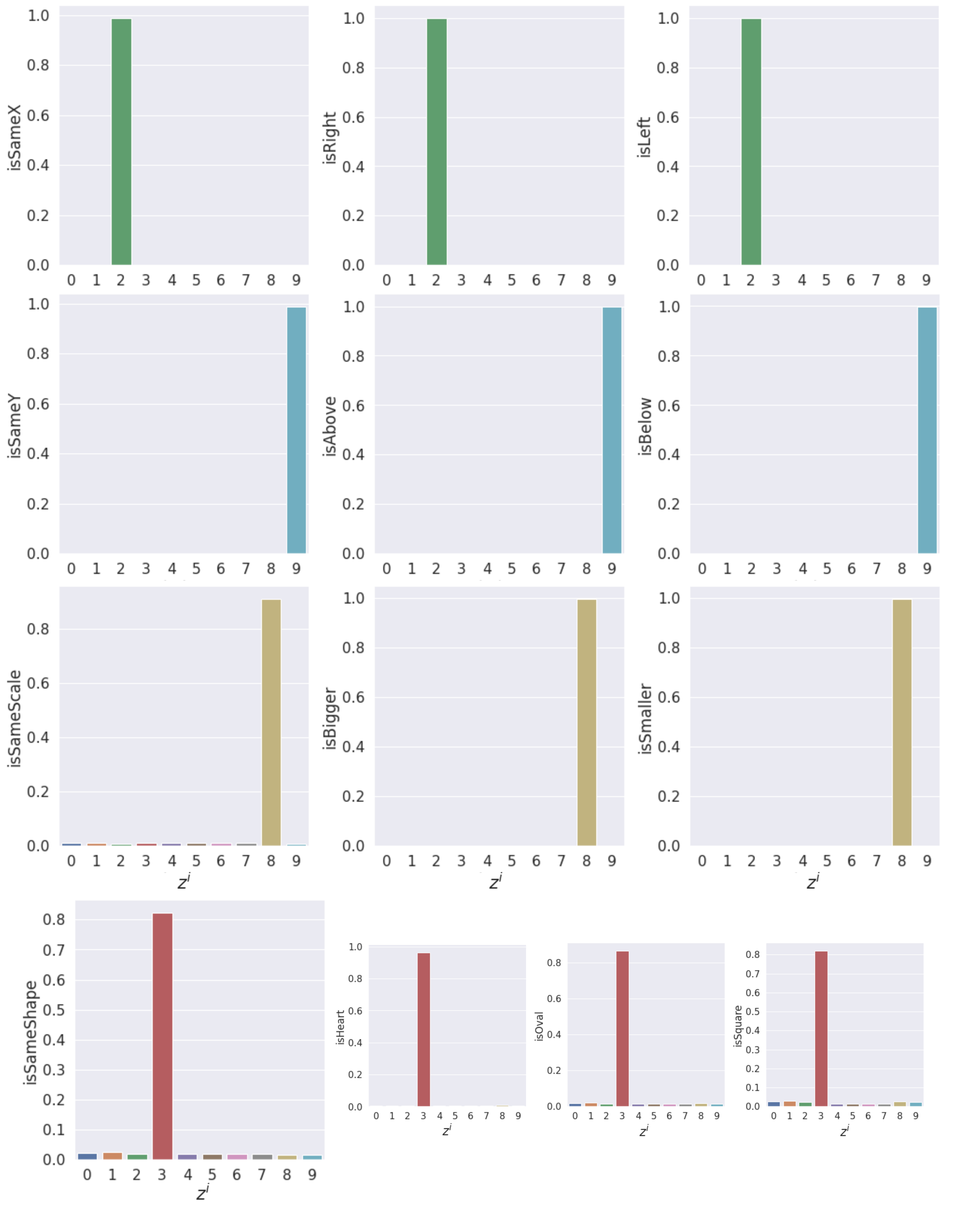}
    \caption{Example of the $\bm{u} = \texttt{Softmax}(\bm{U})$ masks learned by our LFM on \textit{dSprites}. y-axis labels give the relations trained.
    All `same' and categorical relations used $a_0 = 1, a_1 = 0$ with $b_\dagger = 0$, meaning they learn a symmetric transitive relations as expected. Comparative relations on the other hand used $a_0 = 0, a_1 = 1$, thereby learning a transitive asymmetrical relation.}
    \label{fig:MASK_visual_dsprites}
\end{figure}

Finally, to evaluate how well DC can learn a common mask between all semantically related relations, when we include $K>1$ ground truth factors, Figure \ref{fig:MASK_visual_dsprites} shows the learned masks over a set of factors for \textsf{dSprites}. We can clearly see that the latent space is divided between each factor, such that each semantically related relation-decoder calculates any truth-values using only the corresponding subspace. \textbf{DC} relation-decoders are trained for binary relations: \{\textsf{isSameX, isRight, isLeft, isSameY, isAbove, isBelow, isSameScale, isBigger, isSmaller, isSameShape}\} and unary relations \{\textsf{isHeart, isOval, isSquare}\}. Unary relations are learned by setting $\bm{z}_j = \bm{0}$ and $\bm{a}=[1\quad0]$. We indeed see that each set of semantically similar set of relations learn similar masks.

\subsection{Limitations of DC with respect to symbolic attributes of relations}
\begin{table}
  \caption{Relations relevant to the `numeric' semantic factor associated with \textit{MNIST} and their symbolic properties. Other than recursion, all are covered by the proposed relation-decoder.}
  \label{tab:relations_symbolic_properties}
  \centering
        \footnotesize
        \begin{tabular} {l|c|c}
        \toprule
         \addlinespace[0.15em]
         \textbf{Relation} & \textbf{Arity} & \textbf{Symbolic Attributes}\\
         \addlinespace[0.15em]
         \midrule
         $\textsf{isZero}(x_i), \ldots, \textsf{isNine}(x_i)$ & unary & classifier\\
         $\textsf{isEqual}(x_i, x_j)$ & binary & symmetrical, transitive\\
         $\textsf{isGreater}(x_i, x_j), \textsf{isLess}(x_i, x_j)$ & binary & asymmetrical, transitive \\
         $\textsf{isSuccessor}(x_i, x_j), \textsf{isPredecessor}(x_i, x_j)$ & binary & asymmetrical, non-transitive\\
         \addlinespace[0.15em]
         \hdashline[2pt/5pt]
         \addlinespace[0.15em]
         $\textsf{isEven}(x_i), \textsf{isOdd}(x_i)$ & unary & recursively defined\\
         \addlinespace[0.15em]
        \bottomrule
    \end{tabular}
 \end{table}
 
As shown by Table \ref{tab:relations_symbolic_properties}, different relations can have different symbolic properties. This is important to consider when defining an LFM relation-decoder, since each relation-decoder will define a region of input-space which outputs true/false \cite{Gutierrez-Basulto2018-FKG} - we require that these various geometric spaces, which are themselves the result of the relation-decoder parameterization, can accommodate the different symbolic attributes. The proposed Dynamic Comparator relation-decoder (Eqn. \ref{eq:lfm*}) can accommodate the majority of Table \ref{tab:relations_symbolic_properties}, where for simple unary attribute relations, we set $\bm{z}_j = \bm{0}$. However, it cannot model recursively defined relations such as,
$$
 \textsf{isOdd}(a) \implies \textsf{isPredecessor}(b, a) \land \textsf{isEven}(b).
$$
Whilst this can be learned as a true false step function on a separate latent dimension (using $f^\ddagger$), we cannot learn both \textsf{isEven}/\textsf{isOdd} such that they share a common latent subspace with \textsf{isEqual}. This is a core motivation of \textbf{DC}. We have attempted to use periodic function components to model recursively defined relations, as this could accommodate same-dimension encoding, but in practise training became unstable.

\subsection{Latent space structure comparison with NTN+}
\label{sec:apdx:latent_structure_investigations}
\begin{figure}
    \centering
    \includegraphics[width=1.\textwidth]{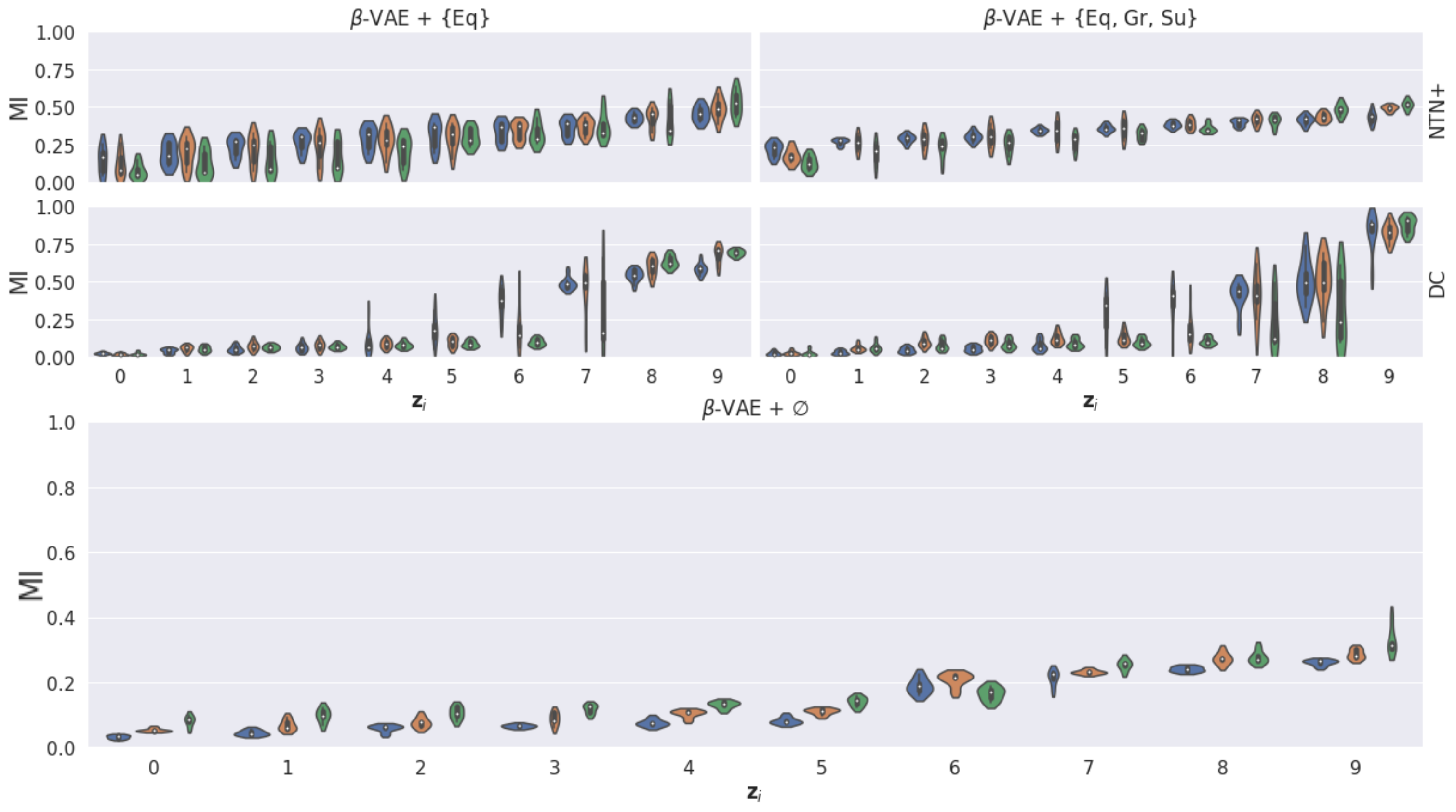}
    \caption{Violin plots showing the normalised mutual information scores for each latent dimension $z^i$ ($i\in 0,\ldots,9$) \wrt the \textit{MNIST} `digit' factor. Results have been ordered to improve objective clarity over the common spectra that each context and relation-decoder setting induces. We report results for three $\beta$ settings indicated by their colour codings as follows: 4-blue, 8-orange and 12-green. }
    \label{fig:mig_scores}
\end{figure}

 \begin{table}
  \caption{Mean and standard deviation MIG scores reported for each relation-decoder and context pairing. Results are included for three $\beta$ settings: 4, 8 and 12.}
  \label{tab:mig_scores}
  \centering
    \footnotesize
    \begin{tabular} {lc|c|c|c|c}
        \toprule
         \addlinespace[0.15em]
         $f_r$ & $\beta$ & $\beta$-VAE + $\emptyset$ & $\beta$-VAE + Eq & $\beta$-VAE + Eq, Gr & $\beta$-VAE + Eq, Gr, Su \\
         \addlinespace[0.15em]
         \midrule
         \multirow{3}{4em}{\textbf{-}} & 4 & 0.02(0.01) & - & - & - \\
            & 8 & 0.02(0.01) & - & - & - \\
            & 12 & 0.04(0.02) & - & - & - \\
        \midrule
        \multirow{3}{4em}{\textbf{NTN+}}& 4 & - & 0.04(0.03) & 0.03(0.03) & 0.02(0.01) \\
            & 8 & - & 0.02(0.02) & 0.02(0.01) & 0.04(0.02) \\
            & 12 & - & 0.08(0.01) & 0.11(0.02) & 0.03(0.01) \\
        \midrule
        \multirow{3}{4em}{\textbf{DC}} & 4 & - & 0.05(0.02) & 0.15(0.18) & 0.31(0.26) \\
            & 8 & - & 0.07(0.05) & 0.21(0.07) & 0.37(0.06) \\
            & 12 & -& 0.08(0.02) & 0.2(0.11) & 0.53(0.26) \\
         \addlinespace[0.15em]
        \bottomrule
    \end{tabular}
 \end{table}
This section exemplifies the latent embedding structure changes induced when using each relation-decoder.
Firstly, we present MIG scores on \textit{MNIST} for each relation-decoder and, for further perspective, when no semi-supervision is used. Table \ref{tab:mig_scores} shows the MIG scores for each context and relation-decoder pairing, when using different values of $\beta$ and Figure \ref{fig:mig_scores} presents the normalised mutual information of each latent dimension \wrt the digit factor. We include the latter as this provides more insight into how the ground truth factor is being encoded in the latent space. 
Looking at both of these results, some key observations are: 
1. Overall, DC achieves the highest MIG scores and shows the greatest disparity of digit factor information being encoded by each latent dimension. 
2. as a general rule, increasing $\beta$ seems to positively influence MIG scores, this is especially the case for DC, where it seems to regularise out digit factor information from all but one dimension at the extreme case;
and 3. although there is no great difference in the average MIG scores between NTN+ semi-supervision and no supervision, we do observe a per-dimension increase in digit factor information being encoded across the latent dimensions.
In summary, we observe a marked increase in digit factor information being encoded by relation-decoders when included versus when performing fully unsupervised representation learning. This explains, at least partly, the significant inductive transfer performance increase observed when adding in semi-supervision. However, it is clear that DC is able to extract and disentangle the digit factor information into the fewest latent dimensions, particularly as more context is included for semi-supervision. This is in contrast to the inductive transfer performance, where we observed an evident decrease in performance, but, interestingly, transductive transfer was observed to improve.
It therefore seems that this increased regularity \wrt the digit factor is beneficial for transductive performance, but not for inductive transfer.

\subsection{Further latent space investigations}
\begin{figure}
    \centering
    \includegraphics[height=.9\textheight]{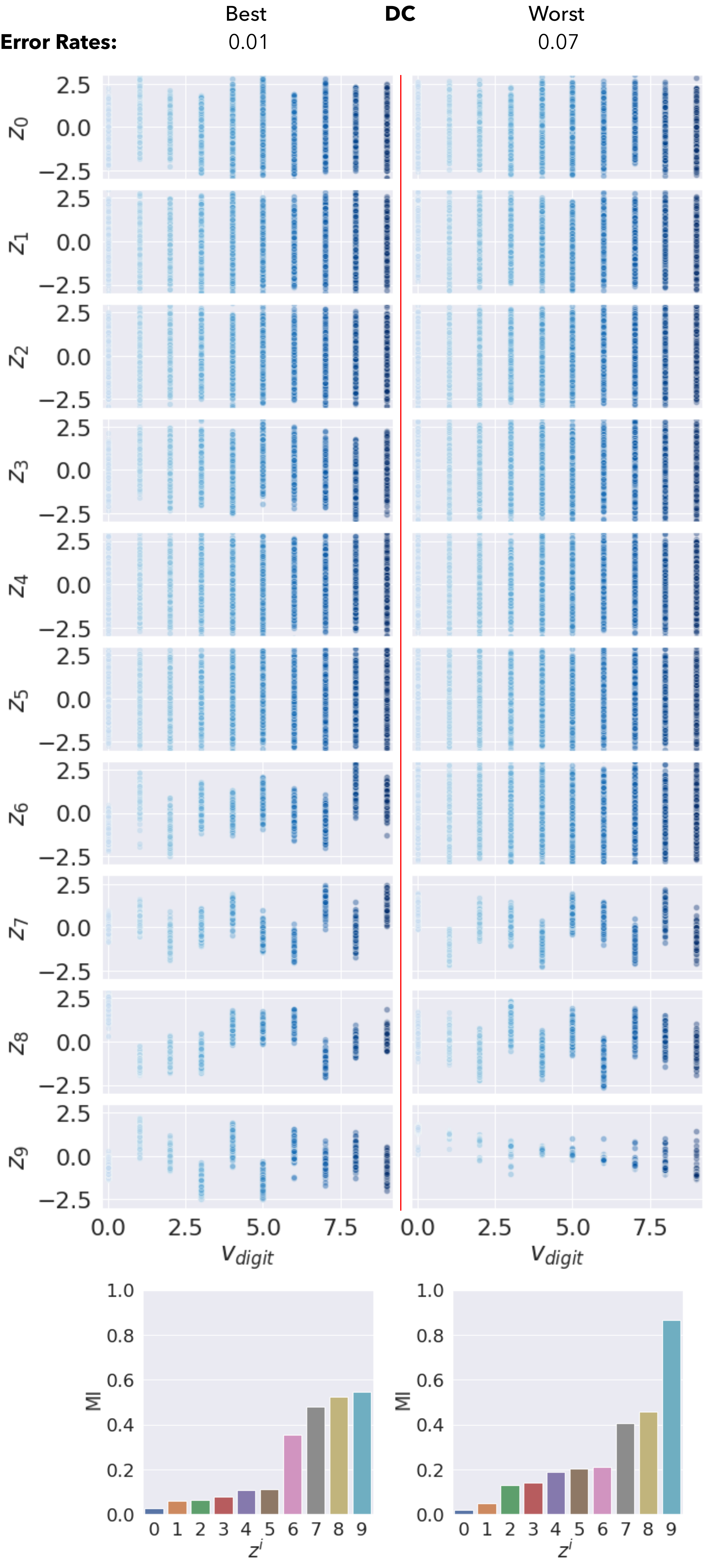}
    \caption{Latent dimension versus digit factor visualisations (top) and normalised mutual information (bottom) when applying semi-supervision with a DC relation-decoder, presented for the best (left) and worse (right) case inductive transfer experiments. Numeric performance values are given in the top row.}
    \label{fig:best_vs_worst_inductive_DC}
\end{figure}

\begin{figure}
    \centering
    \includegraphics[height=.9\textheight]{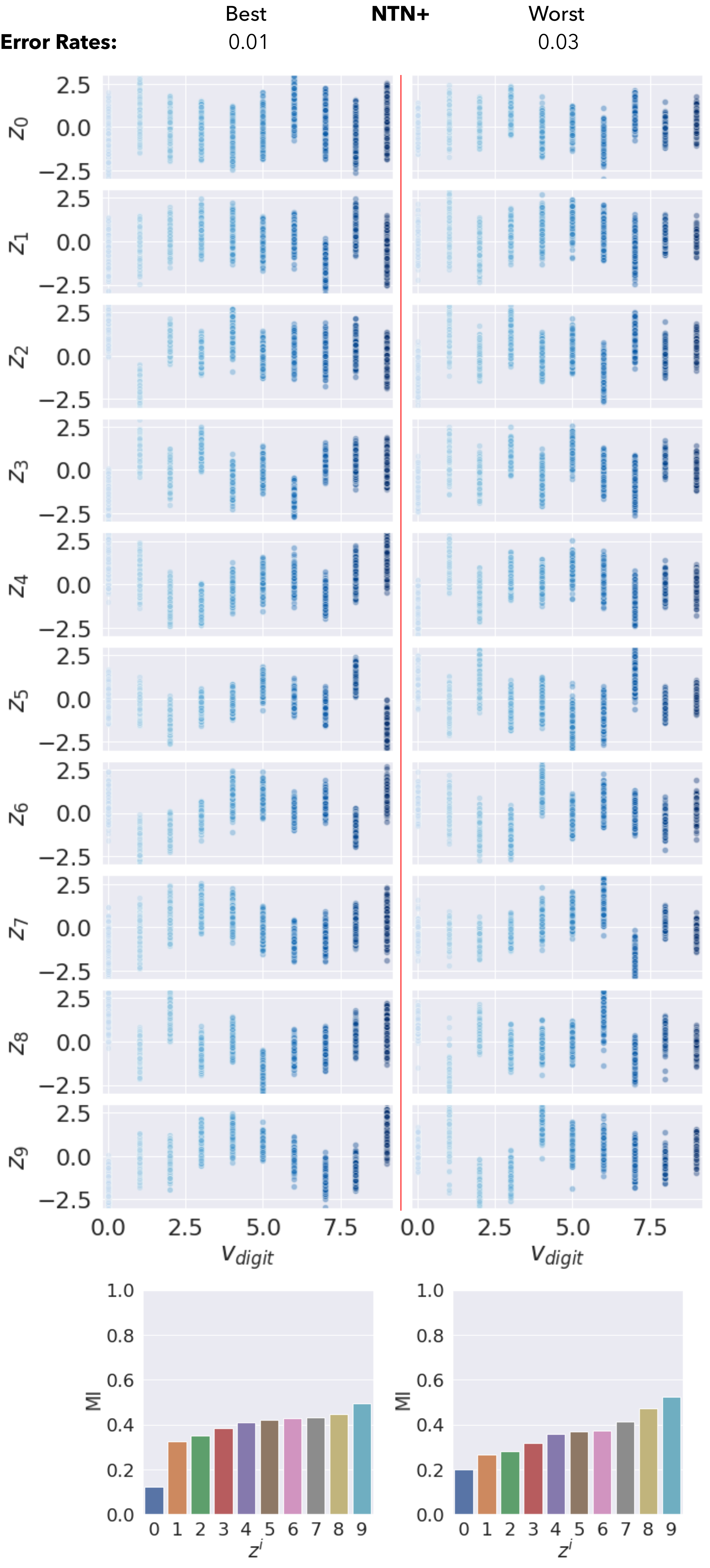}
    \caption{Latent dimension versus digit factor visualisations (top) and normalised mutual information (bottom) when applying semi-supervision with a NTN+ relation-decoder,, presented for the best (left) and worse (right) case inductive transfer experiments. Numeric performance values are given in the top row.}
    \label{fig:best_vs_worst_inductive_NTN}
\end{figure}

In this section we provide further visualisations to substantiate any claims regarding the levels of semantic `structure' that each relation-decoder induces when employed for $\beta$-VAE semi-supervision. These results aid in understanding the interplay between semantic structure and representation transferability.
Figures \ref{fig:best_vs_worst_inductive_DC} and \ref{fig:best_vs_worst_inductive_NTN} present latent dimension versus ground truth digit class scatter plots, for the best/worst performing DC and NTN+ relation-decoder experiments, respectively. 

In each case, the mutual information `spectra' across the latent dimensions are also included to demonstrate how digit class disentanglement affects transfer performance.
In summary, we again see that the improved disentanglement of the digit factor obtained by DC positively improves transductive transfer, but has negatively affects on inductive transfer. 

\begin{figure}
    \centering
    \includegraphics[height=.9\textheight]{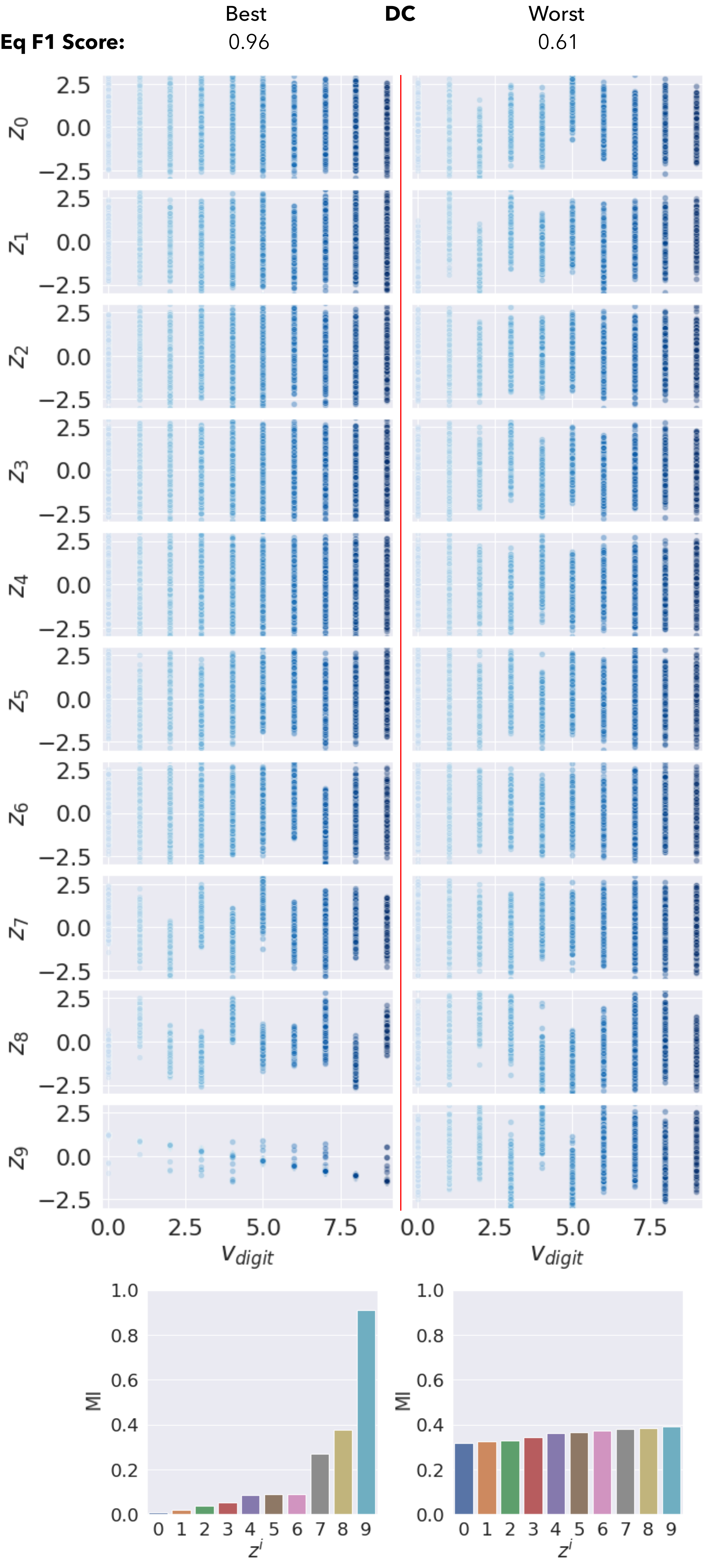}
    \caption{Latent dimension versus digit factor visualisations (top) and normalised mutual information (bottom) when applying semi-supervision with a DC relation-decoder, presented for the best (left) and worse (right) case transductive transfer experiments. Numeric performance values are given in the top row.}
    \label{fig:best_vs_worst_transductive_DC}
\end{figure}
\begin{figure}
    \centering
    \includegraphics[height=.9\textheight]{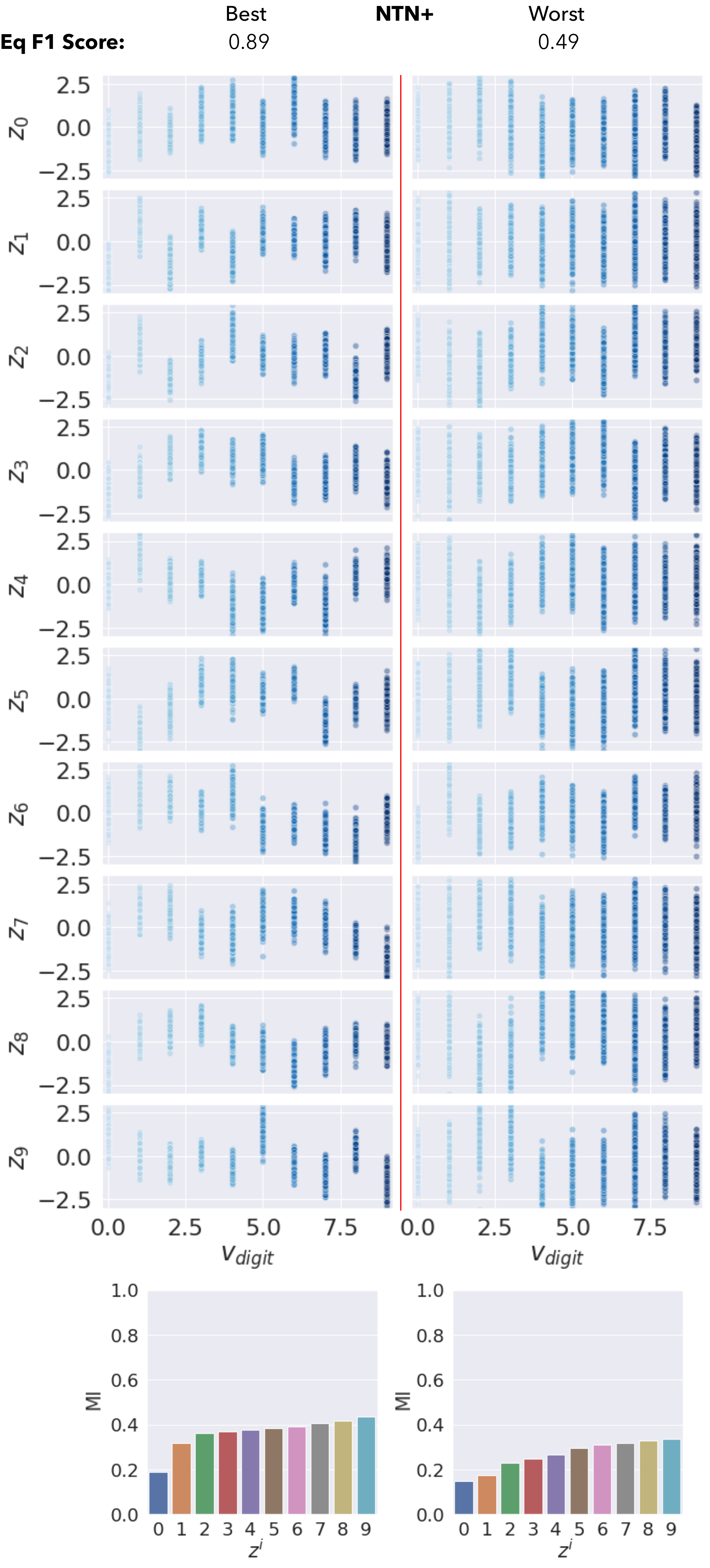}
    \caption{Latent dimension versus digit factor visualisations (top) and normalised mutual information (bottom) when applying semi-supervision with a NTN+ relation-decoder, presented for the best (left) and worse (right) case transductive transfer experiments. Numeric performance values are given in the top row.}
    \label{fig:best_vs_worst_transductive_NTN}
\end{figure}

\begin{figure}
    \centering
    \includegraphics[width=1.\textwidth]{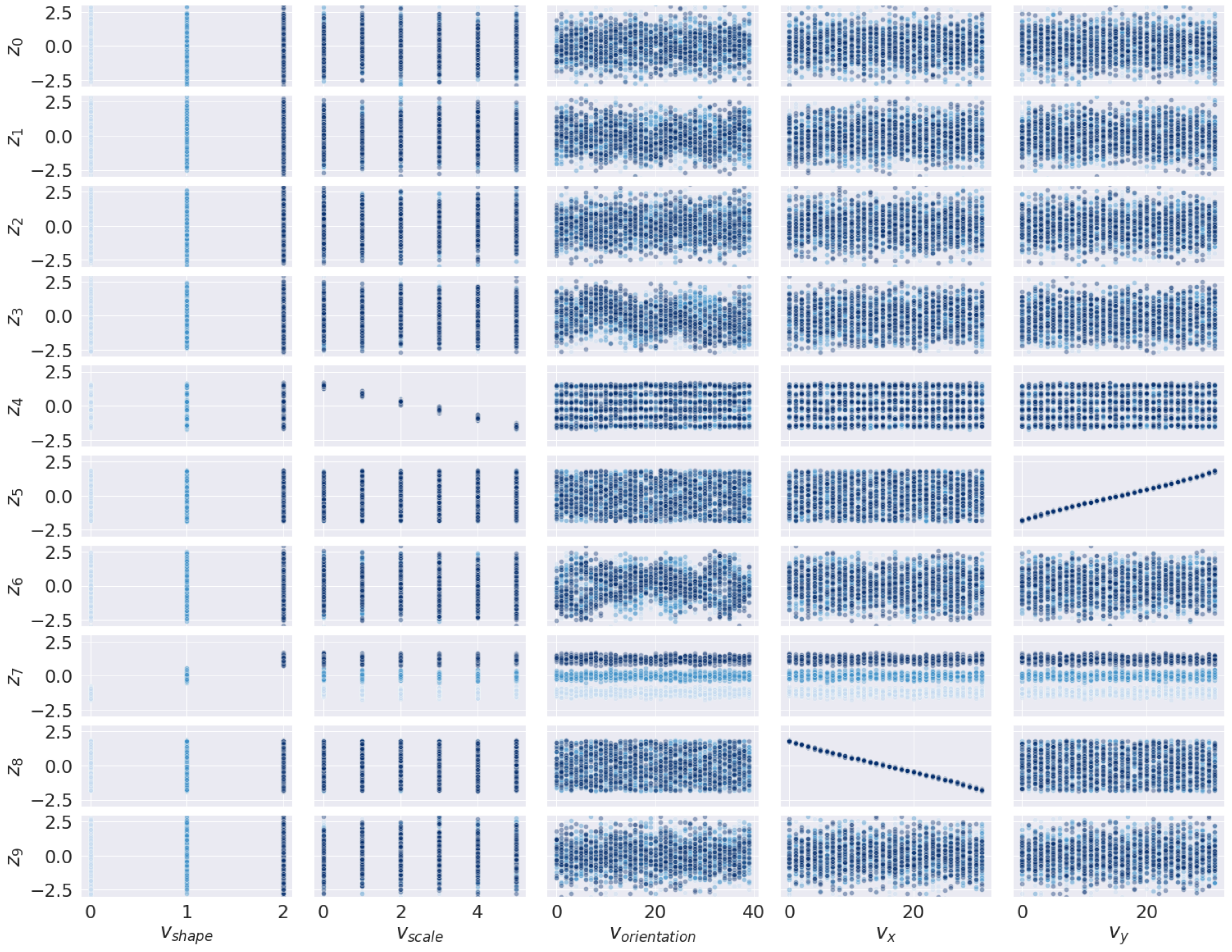}
    \caption{Latent dimension versus ground truth factor visualisations for \textit{dSprites}, when using a DC relation-decoder and training on each relation shown by Figure \ref{fig:MASK_visual_dsprites}.}
    \label{fig:dsprites_full_view_dc}
\end{figure}

\begin{figure}
    \centering
    \includegraphics[width=1.\textwidth]{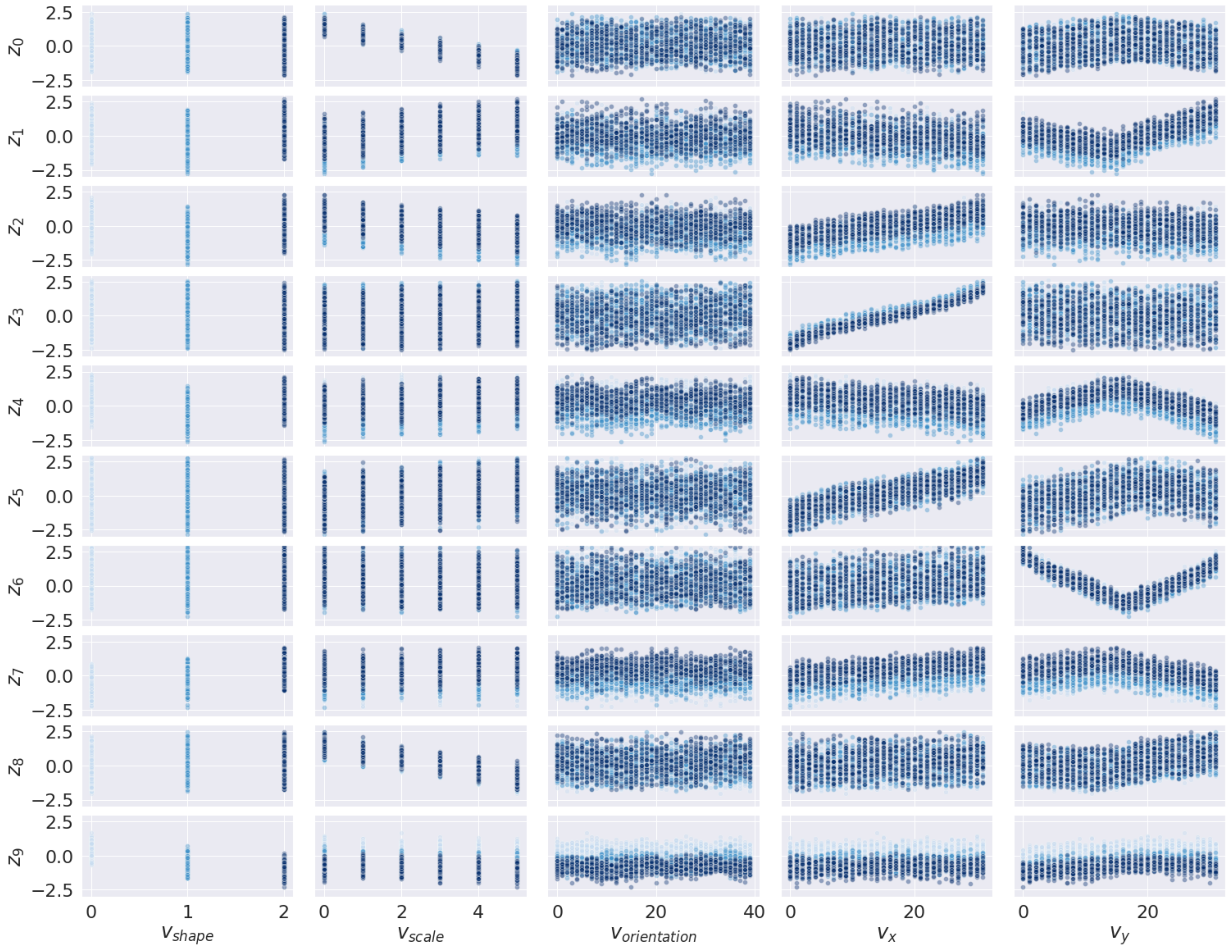}
    \caption{Latent dimension versus ground truth factor visualisations for \textit{dSprites}, when using a NTN+ relation-decoder and training on each relation shown by Figure \ref{fig:MASK_visual_dsprites}.}
    \label{fig:dsprites_full_view_ntn}
\end{figure}

For completeness, Figures \ref{fig:dsprites_full_view_dc} and \ref{fig:dsprites_full_view_ntn} present visualisations of each latent dimension \wrt each ground truth factor, on \textit{dSprites} when training on same relations as in Figure \ref{fig:MASK_visual_dsprites}. We can see that DC produces clearer correlations with each ground truth factor than NTN+.

\newpage

\begin{figure}[H]
    \centering
    \includegraphics[width=0.75\textwidth]{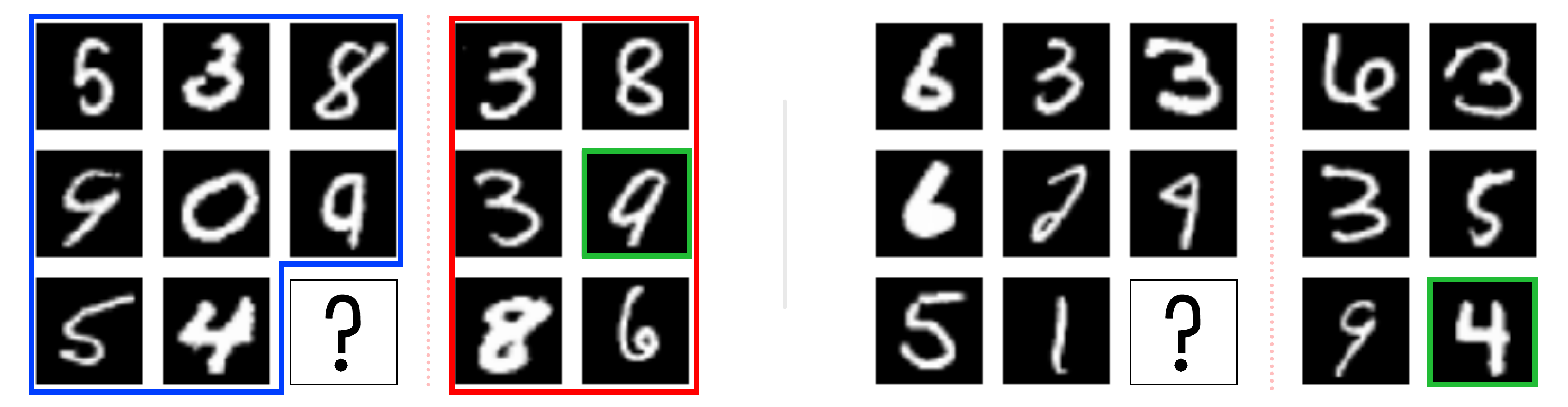}
    \caption{Two examples of our RPM-like \textit{MNIST}-arithmetic tasks, where a reasoner must identify and perform addition (left) or subtraction (right) over \textit{MNIST} digits. Each RPM instance is constructed by a set of tiles corresponding to the context (blue), question (\textbf{?}), answer set (red) and answer (green)}
    \label{fig:pgm_eg}
\end{figure}
\section{Abstract Reasoning over Non-Visual Semantics}
\label{apdx:sec:avr}

This section provides further details regarding the inductive task setup used in the main text.

Recently, abstract reasoning tasks inspired by Raven's Progressive Matrices (RPM) - a well-established measure of non-verbal intelligence \cite{Raven1941-SOP} - have been used to illustrate the generalisation capability of disentangled representations \cite{Steenbrugge2018-IGA, Steenkiste2019-ADR}. Neural networks designed specifically for the RPM task were used in \citet{Barrett2018-MAR}.
In an RPM task, the learner is presented with a panel made of sequences of context tiles, following one or more relational consistencies, and a final question tile. The learner is tasked with identifying the underlying consistency patterns by selecting the missing question tile from a set of possible answers. 
This testing mechanism is useful here as it tests the learner's ability to identify semantic relationships in the data \cite{Kemp2008-TDO} and it is thought that a disentangled representation should express clearly the relevant relationships upon which the reasoning behind the RPM task is to be constructed.
We construct an RPM-inspired task based on the recent work using \textit{MNIST} data for arithmetic \cite{Madsen2020-NAU, Trask2018-NAL}. For each RPM panel, the downstream learner is required first to identify the mathematical operation being performed, and then to apply it to the final panel in order to select the correct answer. See Figure \ref{fig:pgm_eg} for an example.

\textbf{Reasoning Model:} In order to compute the downstream reasoning task, we use a ``Wild Relational Network'' (WReN), a purpose-built architecture designed for RPM tasks \cite{Barrett2018-MAR, Santoro2017-ASN}. 
A WReN leverages a previous relational architecture \cite{Santoro2017-ASN} in order to compute pairwise interaction embeddings for each context-to-context tile pairing and context-to-answer. The model uses two shared neural networks, $g_\theta$ and $f_\phi$, one for interaction learning and another for overall scoring, which ensures that the same reasoning method is performed for each possible answer. 
In the standard WReN, tile image embeddings are acquired by way of a CNN feature extraction module. However, it was found that a VAE-obtained disentangled representation can lead to improvements, for example on sample complexity \cite{Steenbrugge2018-IGA, Steenkiste2019-ADR}.
In this paper, we follow the same procedure: we compare representations obtained from a VAE trained with and without a semi-supervision on our \textit{MNIST} RPM task evaluated using a WReN downstream reasoner but with fewer parameters than was used in \cite{Steenkiste2019-ADR}. 
WReN model details are provided in Section \ref{sec:apdx:model_details}.

\newpage

\section{Background Theory}
\label{apdx:sec:background_theory}

\textbf{Variational AutoEncoders (VAE) -} 
The VAE is derived by introducing an approximate posterior $q_{\bm{\phi}}(\bm{Z}|\bm{X})$, from which a lower bound (commonly referred to as the Evidence LOwer  Bound (ELBO)) on the true marginal $\ln p_{\bm{\theta}}(\bm{X})$ can be obtained by using Jensen's inequality \cite{Kingma2014-AEV}. The VAE maximises the log-probability by maximising this lower bound, given by:
\begin{equation}
    \mathcal{L}_{\beta\text{-VAE}}^{ELBO} = 
\E{q_{\bm{\phi}}(\bm{Z}|\bm{X})}{\log p_{\bm{\theta}}(\bm{X}|\bm{Z})} 
- \beta D_{KL}(q_{\bm{\phi}}(\bm{Z}|\bm{X}) \| p_{\bm{\theta}}(\bm{Z})),
\label{eq:elbo}
\end{equation}
where $q_{\bm{\phi}}(\bm{Z}|\bm{X})$ is the approximate posterior, typically modelled as a neural network encoder with parameters $\bm{\phi}$. Similarly $p_{\bm{\theta}}(\bm{X}|\bm{Z})$ is modelled as a decoder with parameters $\bm{\theta}$ and is calculated as a Monte Carlo estimation. A reparameterization trick is used to enable differentiation through this term (see \cite{Kingma2014-AEV}).
In the $\beta$-VAE \cite{Higgins2017-BVL, Burgess2017-UDI}, an additional $\beta$ scalar hyperparameter was added as it was found to influence disentanglement through stronger distribution matching pressure with the isotropic zero-mean Gaussian  prior $p_{\bm{\theta}}(\bm{Z})$. When $\beta=1$ we obtain the standard VAE objective \cite{Kingma2014-AEV}.

\textbf{Latent Factor Models (LFM) -} 
LFMs are a technique for knowledge graph embedding, where latent representations for data are learned by jointly optimising for them alongside parameterized relation-decoders. These methods are often applied to link prediction, where data that hold similar relations will have similar latent representations, and thus computing held out relations on the entities should produce the correct true/false scoring (see \cite{Trouillon2019-OIA, Nickel2016-ARO, Wang2017-KGE} for further details).
The importance of LFMs is that the relation-decoders parameters and latent representations together provide the semantics. However, unlike in disentanglement, attention is rarely given to the semantic value of each dimension of the latent space.

\textbf{NTN+ -} In this paper, we use a modified Neural Tensor Network (NTN) back-end \cite{Socher2013-RWN} given by:
\begin{align}
    f_{r}(\bm{z}_0, \ldots, \bm{z}_n) &= \sigma(f_{r}^{\text{NTN+}}(\bm{z}')), \label{eq:ntn+} \\
    f_{r}^{\text{NTN+}}(\bm{z}') &= \bm{u}_{r}^\top [\tanh(\bm{z}'^\top \bm{M}_{r} \bm{z}' + \bm{V}_{r} \bm{z}_c + \bm{b}_{r})] \nonumber \\
    \bm{z}_c \in \Real^{n m}, \bm{M}_r \in \mathbb{R}^{k \times n m \times n m} ; &\bm{V}_r \in \mathbb{R}^{k \times n m} ; \bm{b}_r \in \mathbb{R}^{k}, \bm{u}_r \in \mathbb{R}^{k}, \nonumber 
\end{align}
where, $\sigma$ is a sigmoid function used to bound the output of $f^{NTN+}_{r}$ to $[0,1]$ (interpreted as the truth-value of a predicate in a many-valued logic) and $\bm{z}' = (\bm{z}_0;\cdots;\bm{z}_n)$ is a concatenation of the relation-decoder's arguments with $m$-dimensional latent embeddings $\bm{z}_0,\cdots,\bm{z}_n \in \real^m$. The original NTN does not apply the sigmoid nor the concatenation operation, since it was strictly defined for binary relations, whereas NTN+ can accommodate $n$-ary relations and can model boolean relations without requiring any additional arbitrary true/false thresholding.
The only hyperparameter to consider is $k$ which controls the model's capacity \cite{Socher2013-RWN} - in all experiments, we set this to 1.

\newpage

\section{Model Details}
\label{sec:apdx:model_details}
Our experiments were implemented using PyTorch \cite{pytorch}. 
For all models, we use an Adam optimiser with the same parameters for all relation-decoders and VAE models. These are: learning rate of 0.0001, $\text{betas }=(0.9, 0.999)$, $\epsilon = 1 \times 10^{-8}$. No weight decay is used.
In all experiments, we repeat hyperparameter configurations with 5 restarts. Furthermore, for \textit{MNIST}, all datasets are pre-sampled and shared across experiments. These include a $60,000:10,000$ train and test \textit{MNIST} data split, with corresponding knowledge graphs and RPM datasets.
For any \textit{dSprites} experiments, we randomly produce a 8:2 train-test split and produce triples from combinations of the inputs in each sampled image batch.
In all experiments (both \textit{MNIST} and \textit{dSprites}), we use an image batch size of 64.

\subsection{VAE configuration}
In all representation learning experiments, we use a $\beta$-VAE trained for 300,000 steps, following accepted practise from \cite{Locatello2019-CCA, Steenbrugge2018-IGA}.
The encoder-decoder model parameters are given in Table \ref{tab:bvae_model_params} - we include the model configurations used for both \textit{MNIST} and \textit{dSprites} datasets.

\begin{table}[h]
    \caption{Specification of our $\beta$-VAE encoder and decoder model parameters, for both 28$\times$28 (top) and 64$\times$64 (bottom) size input data. I: Input channels, O: Output channels, K: Kernel size, S: Stride, P: Padding, A: Activation}
    \centering
    % 28 dim dSprites model
    \begin{tabular}{l}
        \toprule
        \textbf{Encoder} \\
         Input: $ 28 \times 28 \times N_C$ \\
         \midrule
         \textbf{Layer\_ID ; I ; O ; K ; S ; P ; A} \\
         Conv2d\_1 ; $N_C$  ; 32 ; $4\times4$ ; 2 ; 1 ; ReLU\\
         Conv2d\_2 ; 32 ; 32 ; $4\times4$ ; 2 ; 1 ; ReLU \\
         Conv2d\_3 ; 32 ; 64 ; $3\times3$ ; 2 ; 1 ; ReLU \\
         Conv2d\_4 ; 64 ; 64 ; $2\times2$ ; 2 ; 1 ; ReLU \\
         \midrule
         \textbf{Layer\_ID ; Num Nodes : In - Out ; A} \\
         FC\_z ; 576 - 144 ; ReLU \\
         FC\_z\_mu ; 144 - 10 ; None  \\
         FC\_z\_logvar ; 144 - 10 ; None \\
         \bottomrule
    \end{tabular}
    \hspace{0.5em}
    \begin{tabular}{l}
        \toprule
        \textbf{Decoder} \\
         Input: $ \real^{10} $ \\
         \midrule
         \textbf{Layer\_ID ; Num Nodes : In - Out ; A} \\
         FC\_z ; 10 - 144 ; ReLU \\
         FC\_z\_mu ; 144 - 576 ; ReLU  \\
         \midrule
         \textbf{Layer\_ID ; I ; O ; K ; S ; P ; A} \\
         UpConv2d\_1 ; 64 ; 64 ; $2\times2$ ; 2 ; 1 ; ReLU \\
         UpConv2d\_1 ; 64 ; 32 ; $3\times3$ ; 2 ; 1 ; ReLU \\
         UpConv2d\_1 ; 32 ; 32 ; $4\times4$ ; 2 ; 1 ; ReLU \\
         UpConv2d\_1 ; 32 ; $N_C$ ; $4\times4$ ; 2 ; 1 ; Sigmoid \\
         \\
         \bottomrule
    \end{tabular}
    \newline
    \vspace*{0.5 cm}
    \newline
    % 64 dim dSprites model
    \begin{tabular}{l}
        \toprule
        \textbf{Encoder} \\
         Input: $ 64 \times 64 \times N_C$ \\
         \midrule
         \textbf{Layer\_ID ; I ; O ; K ; S ; P ; A} \\
         Conv2d\_1 ; $N_C$  ; 32 ; $4\times4$ ; 2 ; 1 ; ReLU\\
         Conv2d\_2 ; 32 ; 32 ; $4\times4$ ; 2 ; 1 ; ReLU \\
         Conv2d\_3 ; 32 ; 64 ; $4\times4$ ; 2 ; 1 ; ReLU \\
         Conv2d\_4 ; 64 ; 64 ; $4\times4$ ; 2 ; 1 ; ReLU \\
         \midrule
         \textbf{Layer\_ID ; Num Nodes : In - Out ; A} \\
         FC\_z ; 1024 - 256 ; ReLU \\
         FC\_z\_mu ; 256 - 10 ; None  \\
         FC\_z\_logvar ; 256 - 10 ; None \\
         \bottomrule
    \end{tabular}
    \hspace{0.5em}
    \begin{tabular}{l}
        \toprule
        \textbf{Decoder} \\
         Input: $ \real^{10} $ \\
         \midrule
         \textbf{Layer\_ID ; Num Nodes : In - Out ; A} \\
         FC\_z ; 10 - 256 ; ReLU \\
         FC\_z\_mu ; 256 - 1024 ; ReLU  \\
         \midrule
         \textbf{Layer\_ID ; I ; O ; K ; S ; P ; A} \\
         UpConv2d\_1 ; 64 ; 64 ; $4\times4$ ; 2 ; 1 ; ReLU \\
         UpConv2d\_1 ; 64 ; 32 ; $4\times4$ ; 2 ; 1 ; ReLU \\
         UpConv2d\_1 ; 32 ; 32 ; $4\times4$ ; 2 ; 1 ; ReLU \\
         UpConv2d\_1 ; 32 ; $N_C$ ; $4\times4$ ; 2 ; 1 ; Sigmoid \\
         \\
         \bottomrule
    \end{tabular}
    \label{tab:bvae_model_params}
\end{table}

\subsection{WReN module}
For abstract reasoning tasks, we follow the setup of \cite{Steenbrugge2018-IGA}, training for 100,000 steps and a batch size of 32, and testing 100 RPM new samples after each 1000 steps. As in \cite{Steenbrugge2018-IGA}, we ensure that each RPM instance is new, meaning that the training set consists of 3.2$\times 10^6$ samples.
In this paper, each RPM panel consists of eight context tiles $C = \{c_1, \ldots, c_8 \}$ with six possible answer tiles $A = \{a_1, \ldots, a_6 \}$. For each possible answer, the WReN reasoning module computes: 
\begin{align}
    WReN(a_k, C) &= f_\phi \big(\sum_{\bm{z}_i, \bm{z}_j \in Z} g_\theta(h_\gamma (\bm{z}_i), h_\gamma(\bm{z}_j))\big), \\ 
    Z = \{ \psi(c_1), \ldots &\psi(c_8) \cup \psi(a_k)\}, \quad \psi(\cdot) = CNN(\cdot)\ (\lor\ \psi_{enc}(\cdot)),
\end{align}
where $f_\theta$ is a scoring multilayer perceptron (MLP), which takes as input an aggregation over inter-tile interactions, as computed by the relation network function $g_\theta$.
As in \cite{Steenbrugge2018-IGA, Steenkiste2019-ADR}, instead of using a CNN feature extractor as in the original model, we replace the initial CNN feature extractor \cite{Barrett2018-MAR} with pre-trained representations (\ie $\bm{z}_i, \bm{z}_j$) taken from the VAE bottleneck as extracted by the VAE encoder: $\psi_{enc}$.
Finally, $h_\gamma$ serves the purpose of incorporating positional features into each tile, where each tile has its feature vector concatenated with a one-hot position vector. This is then passed through a single-layer fully connected MLP to obtain each tiles' feature vector. Note that all answer tiles are considered to be at the final position (position 9) of the panel.
See \cite{Barrett2018-MAR, Santoro2017-ASN} for more information on the overall WReN architecture. 
The WReN model parameters are provided in Table \ref{tab:wren_model_params}, where the reduced parameters sizes, used in the paper's main text experiments, are given in parenthesis.

\begin{table}[h]
    \caption{Specification of our the WReN model parameters. 
    A: Activation}
    \centering
    \begin{tabular}{l}
        \toprule
        \textbf{Scoring function} $f_\phi$ \\
         Input: $ 28 \times 28 \times N_C$ \\
         \midrule
        \textbf{ Layer\_ID ; Num Nodes : In - Out ; A} \\
         FC\_1 ; 64 ; 64 ; ReLU \\
         FC\_2 ; 64 ; 64 ; ReLU  \\
         -- Drop-out layer $p=0.5$ -- \\
         FC\_4 ; 64 ; 1 ; None  \\
         \bottomrule
    \end{tabular}
     \hspace{2em}
    \begin{tabular}{l}
        \toprule
        \textbf{Relation net} $g_\theta$ \\
         Input: concatenation $ [\real^{64} ; \real^{64}]$ \\
         \midrule
         \textbf{Layer\_ID ; Num Nodes : In ; Out ; A} \\
         FC\_1 ; 2*64 ; 128 ; ReLU \\
         FC\_2 ; 128 ; 128 ; ReLU  \\
         FC\_3 ; 128 ; 128 ; ReLU  \\
         FC\_4 ; 128 ; 64 ; ReLU  \\
         \bottomrule
    \end{tabular}
    \newline
    \vspace*{0.5 cm}
    \newline
    \begin{tabular}{l}
        \toprule
         \textbf{Tile-position feature projection} $h_\gamma$ \\
         Input: $ \real^{10} $ + $\real^9$ position one-hot vector \\
         \midrule
         \textbf{Layer\_ID ; Num Nodes : In ;  Out ; A} \\
         FC\_1 ; 10+9 ; 64 ; ReLU \\
         \bottomrule
    \end{tabular}
    \label{tab:wren_model_params}
\end{table}


\begin{thebibliography}{35}
\providecommand{\natexlab}[1]{#1}
\providecommand{\url}[1]{#1}
\csname url@samestyle\endcsname
\providecommand{\newblock}{\relax}
\providecommand{\bibinfo}[2]{#2}
\providecommand{\BIBentrySTDinterwordspacing}{\spaceskip=0pt\relax}
\providecommand{\BIBentryALTinterwordstretchfactor}{4}
\providecommand{\BIBentryALTinterwordspacing}{\spaceskip=\fontdimen2\font plus
\BIBentryALTinterwordstretchfactor\fontdimen3\font minus
  \fontdimen4\font\relax}
\providecommand{\BIBforeignlanguage}[2]{{%
\expandafter\ifx\csname l@#1\endcsname\relax
\typeout{** WARNING: IEEEtranN.bst: No hyphenation pattern has been}%
\typeout{** loaded for the language `#1'. Using the pattern for}%
\typeout{** the default language instead.}%
\else
\language=\csname l@#1\endcsname
\fi
#2}}
\providecommand{\BIBdecl}{\relax}
\BIBdecl

\bibitem[Pan and Yang(2010)]{Pan2009-ASO}
S.~J. Pan and Q.~Yang, ``A survey on transfer learning,'' \emph{IEEE Trans. on
  Knowl. and Data Eng.}, vol.~22, no.~10, p. 1345–1359, Oct. 2010.

\bibitem[Bengio et~al.(2013)Bengio, Courville, and Vincent]{Bengio2013-RLR}
Y.~Bengio, A.~Courville, and P.~Vincent, ``{Representation learning: A review
  and new perspectives},'' \emph{IEEE Transactions on Pattern Analysis and
  Machine Intelligence}, vol.~35, no.~8, pp. 1798--1828, 2013.

\bibitem[Higgins et~al.(2017)Higgins, Matthey, Pal, Burgess, Glorot, Botvinick,
  Mohamed, and Lerchner]{Higgins2017-BVL}
I.~Higgins, L.~Matthey, A.~Pal, C.~Burgess, X.~Glorot, M.~Botvinick,
  S.~Mohamed, and A.~Lerchner, ``{beta-VAE: Learning Basic Visual Concepts with
  a Constrained Variational Framework},'' in \emph{5th International Conference
  on Learning Representations, {\{}ICLR{\}}}, Toulon, France, 2017.

\bibitem[Shu et~al.(2020)Shu, Chen, Kumar, Ermon, and Poole]{Shu2020-WSD}
R.~Shu, Y.~Chen, A.~Kumar, S.~Ermon, and B.~Poole, ``{Weakly Supervised
  Disentanglement With Guarantees},'' in \emph{8th International Conference on
  Learning Representations, ICLR}, Addis Ababa, Ethiopia, 2020.

\bibitem[Locatello et~al.(2020)Locatello, Poole, R{\"{a}}tsch, Sch{\"{o}}lkopf,
  Bachem, and Tschannen]{Locatello2020-WSD}
F.~Locatello, B.~Poole, G.~R{\"{a}}tsch, B.~Sch{\"{o}}lkopf, O.~Bachem, and
  M.~Tschannen, ``{Weakly-Supervised Disentanglement Without Compromises},''
  \emph{CoRR}, vol. abs/2002.0, 2020.

\bibitem[Chen and Batmanghelich(2020)]{Chen2019-WSD}
J.~Chen and K.~Batmanghelich, ``{Weakly Supervised Disentanglement by Pairwise
  Similarities},'' in \emph{Proceedings of the 32nd AAAI Conference on
  Artificial Intelligence, AAAI}, New York, NY, USA, 2020.

\bibitem[Karaletsos et~al.(2016)Karaletsos, Belongie, and
  R{\"{a}}tsch]{Karaletsos2016-WCH}
T.~Karaletsos, S.~Belongie, and G.~R{\"{a}}tsch, ``{When crowds hold
  privileges: Bayesian unsupervised representation learning with oracle
  constraints},'' in \emph{4th International Conference on Learning
  Representations, {\{}ICLR{\}}}, San Juan, Puerto Rico, 2016, pp. 1--16.

\bibitem[Kingma et~al.(2014)Kingma, Mohamed, Rezende, and
  Welling]{Kingma2014-SSL}
D.~P. Kingma, S.~Mohamed, D.~J. Rezende, and M.~Welling, ``{Semi-supervised
  Learning with Deep Generative Models},'' in \emph{Advances in Neural
  Information Processing Systems 27: Annual Conference on Neural Information
  Processing Systems}, Montreal, Quebec, Canada, 2014, pp. 3581----3589.

\bibitem[Feng et~al.(2018)Feng, Zeng, Wang, Tao, Ke, and Song]{Feng2018-DSD}
Z.~Feng, A.~Zeng, X.~Wang, D.~Tao, C.~Ke, and M.~Song, ``{Dual swap
  disentangling},'' in \emph{Advances in Neural Information Processing Systems
  32}, Montreal, Canada, 2018, pp. 5894--5904.

\bibitem[Locatello et~al.(2019)Locatello, Bauer, Lucic, R\"atsch, Gelly,
  Sch\"olkopf, and Bachem]{Locatello2019-CCA}
F.~Locatello, S.~Bauer, M.~Lucic, G.~R\"atsch, S.~Gelly, B.~Sch\"olkopf, and
  O.~Bachem, ``{Challenging Common Assumptions in the Unsupervised Learning of
  Disentangled Representations},'' in \emph{Proceedings of the 36th
  International Conference on Machine Learning,{\{}ICML{\}}}, Long Beach,
  California, USA, 2019, pp. 4114----4124.

\bibitem[Chen and Batmanghelich(2019)]{Chen2019-ROV}
J.~Chen and K.~Batmanghelich, ``Robust ordinal {VAE:} employing noisy pairwise
  comparisons for disentanglement,'' \emph{CoRR}, vol. abs/1910.05898, 2019.

\bibitem[Trouillon et~al.(2019)Trouillon, Gaussier, Dance, and
  Bouchard]{Trouillon2019-OIA}
T.~Trouillon, {\'{E}}.~Gaussier, C.~R. Dance, and G.~Bouchard, ``{On inductive
  abilities of latent factor models for relational learning},'' \emph{Journal
  of Artificial Intelligence Research}, vol.~64, pp. 21--53, 2019.

\bibitem[Nickel et~al.(2016)Nickel, Murphy, Tresp, and
  Gabrilovich]{Nickel2016-ARO}
M.~Nickel, K.~Murphy, V.~Tresp, and E.~Gabrilovich, ``{A review of relational
  machine learning for knowledge graphs},'' \emph{Proceedings of the IEEE},
  vol. 104, no.~1, pp. 11--33, 2016.

\bibitem[Wang et~al.(2017)Wang, Mao, Wang, and Guo]{Wang2017-KGE}
Q.~Wang, Z.~Mao, B.~Wang, and L.~Guo, ``{Knowledge graph embedding: A survey of
  approaches and applications},'' \emph{IEEE Transactions on Knowledge and Data
  Engineering}, vol.~29, no.~12, pp. 2724----2743, 2017.

\bibitem[Socher et~al.(2013)Socher, Chen, Manning, Chen, and
  Ng]{Socher2013-RWN}
R.~Socher, D.~Chen, C.~Manning, D.~Chen, and A.~Ng, ``{Reasoning With Neural
  Tensor Networks for Knowledge Base Completion},'' in \emph{Advances in Neural
  Information Processing Systems 26: 27th Annual Conference on Neural
  Information Processing Systems}, 2013, pp. 926--934.

\bibitem[LeCun and Cortes(2010)]{mnist}
\BIBentryALTinterwordspacing
Y.~LeCun and C.~Cortes, ``{MNIST} handwritten digit database,'' 2010. [Online].
  Available: \url{http://yann.lecun.com/exdb/mnist/}
\BIBentrySTDinterwordspacing

\bibitem[Donadello et~al.(2017)Donadello, Serafini, and d'Avila
  Garcez]{Donadello2017-LTN}
I.~Donadello, L.~Serafini, and A.~d'Avila Garcez, ``{Logic Tensor Networks for
  Semantic Image Interpretation},'' in \emph{Proceedings of the Twenty-Sixth
  International Joint Conference on Artificial Intelligence}, 2017, pp.
  1596----1602.

\bibitem[Serafini and Garcez(2016)]{Serafini2016-LTN}
L.~Serafini and A.~D. Garcez, ``{Logic tensor networks: Deep learning and
  logical reasoning from data and knowledge},'' in \emph{Proceedings of the
  11th International Workshop on Neural-Symbolic Learning and Reasoning
  (NeSy'16) co-located with the Joint Multi-Conference on Human-Level
  Artificial Intelligence {\{}(HLAI{\}} 2016)}, New York, NY, USA, 2016.

\bibitem[Chen et~al.(2018)Chen, Li, Grosse, and Duvenaud]{Chen2018-ISD}
R.~T.~Q. Chen, X.~Li, R.~B. Grosse, and D.~Duvenaud, ``{Isolating Sources of
  Disentanglement in Variational Autoencoders},'' in \emph{Advances in Neural
  Information Processing Systems 31: Annual Conference on Neural Information
  Processing Systems}, Montreal, Quebec, Canada, 2018, pp. 2615----2625.

\bibitem[Kumar et~al.(2018)Kumar, Sattigeri, and Balakrishnan]{Kumar2018-VID}
A.~Kumar, P.~Sattigeri, and A.~Balakrishnan, ``{Variational inference of
  disentangled latent concepts from unlabeled observations},'' in \emph{6th
  International Conference on Learning Representations, {\{}ICLR{\}}},
  Vancouver, BC, Canada, 2018.

\bibitem[Kim and Mnih(2018)]{Kim2018-DF}
H.~Kim and A.~Mnih, ``{Disentangling by Factorising},'' in \emph{Proceedings of
  the 35th International Conference on Machine Learning, {\{}ICML{\}}},
  Stockholm, Sweden, 2018, pp. 2654----2663.

\bibitem[Ridgeway and Mozer(2018)]{Ridgeway2018-LDD}
K.~Ridgeway and M.~C. Mozer, ``{Learning Deep Disentangled Embeddings With the
  F-Statistic Loss},'' in \emph{Advances in Neural Information Processing
  Systems 31: Annual Conference on Neural Information Processing Systems},
  Montreal, Quebec, Canada, 2018, pp. 185----194.

\bibitem[Eastwood and Williams(2018)]{Eastwood2018-FQE}
C.~Eastwood and C.~K.~I. Williams, ``{A framework for the quantitative
  evaluation of disentangled representations},'' in \emph{6th International
  Conference on Learning Representations, {\{}ICLR{\}}}, Vancouver, BC, Canada,
  2018.

\bibitem[Burgess et~al.(2017)Burgess, Higgins, Pal, Matthey, Watters,
  Desjardins, and Lerchner]{Burgess2017-UDI}
\BIBentryALTinterwordspacing
C.~P. Burgess, I.~Higgins, A.~Pal, L.~Matthey, N.~Watters, G.~Desjardins, and
  A.~Lerchner, ``{Understanding disentangling in {$\beta$-VAE}},'' in
  \emph{Advances in Neural Information Processing Systems 30}, no. Nips, Long
  Beach, CA, USA, 2017. [Online]. Available:
  \url{http://arxiv.org/abs/1804.03599}
\BIBentrySTDinterwordspacing

\bibitem[Steenbrugge et~al.(2018)Steenbrugge, Leroux, Verbelen, and
  Dhoedt]{Steenbrugge2018-IGA}
X.~Steenbrugge, S.~Leroux, T.~Verbelen, and B.~Dhoedt, ``{Improving
  Generalization for Abstract Reasoning Tasks Using Disentangled Feature
  Representations},'' in \emph{Neural Information Processing Systems (NeurIPS)
  Workshop on Relational Representation Learning}, Montreal, Canada, 2018.

\bibitem[van Steenkiste et~al.(2019)van Steenkiste, Locatello, Schmidhuber, and
  Bachem]{Steenkiste2019-ADR}
S.~van Steenkiste, F.~Locatello, J.~Schmidhuber, and O.~Bachem, ``{Are
  Disentangled Representations Helpful for Abstract Visual Reasoning?}'' in
  \emph{Advances in Neural Information Processing Systems 32: Annual Conference
  on Neural Information Processing Systems}, Vancouver, BC, Canada, 2019, pp.
  14\,222----14\,235.

\bibitem[Barrett et~al.(2018)Barrett, Hill, Santoro, Morcos, and
  Lillicrap]{Barrett2018-MAR}
D.~G. Barrett, F.~Hill, A.~Santoro, A.~S. Morcos, and T.~Lillicrap,
  ``{Measuring abstract reasoning in neural networks},'' \emph{35th
  International Conference on Machine Learning, ICML 2018}, vol.~10, pp.
  7118--7127, 2018.

\bibitem[Guti{\'{e}}rrez-Basulto and
  Schockaert(2018)]{Gutierrez-Basulto2018-FKG}
V.~Guti{\'{e}}rrez-Basulto and S.~Schockaert, ``{From Knowledge Graph Embedding
  to Ontology Embedding? An Analysis of the Compatibility between Vector Space
  Representations and Rules},'' in \emph{Principles of Knowledge Representation
  and Reasoning: Proceedings of the Sixteenth International Conference}, Tempe,
  Arizona, US, 2018.

\bibitem[Raven(1941)]{Raven1941-SOP}
J.~C. Raven, ``Standardization of progressive matrices, 1938,'' \emph{British
  Journal of Medical Psychology}, vol.~19, no.~1, pp. 137--150, 1941.

\bibitem[Kemp and Tenenbaum(2008)]{Kemp2008-TDO}
C.~Kemp and J.~B. Tenenbaum, ``{The discovery of structural form},''
  \emph{Proceedings of the National Academy of Sciences of the United States of
  America}, vol. 105, no.~31, pp. 10\,687--10\,692, 2008.

\bibitem[Madsen and Johansen(2020)]{Madsen2020-NAU}
A.~Madsen and A.~R. Johansen, ``{Neural Arithmetic Units},'' in \emph{8th
  International Conference on Learning Representations, ICLR}, Addis Ababa,
  Ethiopia, 2020.

\bibitem[Trask et~al.(2018)Trask, Hill, Reed, Rae, Dyer, and
  Blunsom]{Trask2018-NAL}
A.~Trask, F.~Hill, S.~E. Reed, J.~W. Rae, C.~Dyer, and P.~Blunsom, ``Neural
  arithmetic logic units,'' in \emph{Advances in Neural Information Processing
  Systems 31}, Montreal, Canada, 2018, pp. 8046--8055.

\bibitem[Santoro et~al.(2017)Santoro, Raposo, Barrett, Malinowski, Pascanu,
  Battaglia, and Lillicrap]{Santoro2017-ASN}
A.~Santoro, D.~Raposo, D.~G. Barrett, M.~Malinowski, R.~Pascanu, P.~Battaglia,
  and T.~Lillicrap, ``{A simple neural network module for relational
  reasoning},'' in \emph{Advances in Neural Information Processing Systems 30},
  Long Beach, CA, USA, 2017.

\bibitem[Kingma and Welling(2014)]{Kingma2014-AEV}
D.~P. Kingma and M.~Welling, ``{Auto-Encoding Variational Bayes},'' in
  \emph{Proceedings of the 2nd International Conference on Learning
  Representations}, Banff, Alberta, Canada, 2014.

\bibitem[Paszke et~al.(2019)Paszke, Gross, Massa, Lerer, Bradbury, Chanan,
  Killeen, Lin, Gimelshein, Antiga, Desmaison, Kopf, Yang, DeVito, Raison,
  Tejani, Chilamkurthy, Steiner, Fang, Bai, and Chintala]{pytorch}
\BIBentryALTinterwordspacing
A.~Paszke, S.~Gross, F.~Massa, A.~Lerer, J.~Bradbury, G.~Chanan, T.~Killeen,
  Z.~Lin, N.~Gimelshein, L.~Antiga, A.~Desmaison, A.~Kopf, E.~Yang, Z.~DeVito,
  M.~Raison, A.~Tejani, S.~Chilamkurthy, B.~Steiner, L.~Fang, J.~Bai, and
  S.~Chintala, ``Pytorch: An imperative style, high-performance deep learning
  library,'' in \emph{Advances in Neural Information Processing Systems 32},
  H.~Wallach, H.~Larochelle, A.~Beygelzimer, F.~d\textquotesingle
  Alch\'{e}-Buc, E.~Fox, and R.~Garnett, Eds.\hskip 1em plus 0.5em minus
  0.4em\relax Curran Associates, Inc., 2019, pp. 8024--8035. [Online].
  Available:
  \url{http://papers.neurips.cc/paper/9015-pytorch-an-imperative-style-high-performance-deep-learning-library.pdf}
\BIBentrySTDinterwordspacing

\end{thebibliography}
\end{document}